\documentclass[10pt,twocolumn,letterpaper]{article}

\usepackage[pagenumbers]{cvpr}

\usepackage[T1]{fontenc}

\usepackage{cuted} 

\usepackage{currfile} 

\usepackage{caption} 

\usepackage{bibunits}
\defaultbibliographystyle{ieeenat_fullname}
\defaultbibliography{main}

\definecolor{cvprblue}{rgb}{0.21,0.49,0.74}
\usepackage[
    breaklinks,
    colorlinks,
    allcolors=cvprblue
]{hyperref}

\usepackage{algorithm}
\usepackage{algpseudocode}

\usepackage{xcolor}
\def\eg{\emph{e.g.}}
\def\etal{\emph{et al. }}
\def\ie{\emph{i.e.}}

\definecolor{hl}{RGB}{235,230,222}

\def\blue#1{\textcolor{cvprblue}{#1}}

\definecolor{dotpink}{HTML}{EFA0C8}
\definecolor{dotgreen}{HTML}{7BCB98}

\usepackage{booktabs}
\usepackage{graphicx}
\usepackage{amssymb}
\usepackage{adjustbox}
\usepackage{multirow}

\title{Iterative Collaborative Source-free Adaptation for Unseen Face Forgery Detection}

\author{
    Hanqing Huang$^{1}$ \quad
    Junlin Lai$^{1}$ \quad
    Keji Song$^{1}$ \quad
    Jiaran Zhou$^{1}$ \quad
    Yanbo Fan$^{2}$ \quad
    Yuezun Li$^{1,}$\thanks{Corresponding author.}\\
    $^{1}$School of Computer Science and Technology, Ocean University of China \\ $^{2}$Nanjing University
}

\begin{document}

\begin{bibunit}[ieeenat_fullname]

\maketitle

\begin{abstract}
This paper investigates a practical problem: adapting multiple source-trained face forgery detection models to detect unseen face forgeries without access to source data. Since forgery traces are far less salient than semantic cues, existing adaptation methods are highly limited. To address this, we propose Iterative Collaborative Source-free Adaptation (ICSA), a multi-source adaptation framework comprising three key components: 1) a Source Model Reliability Estimation strategy that assesses the reliability of each source model, 2) a Collaborative Distillation strategy that distills knowledge based on itself and other models, and 3) an Iterative Adaptation scheme that repeats the above process to progressively refine source models. Comprehensive experiments on the DF40 dataset against seven baseline methods demonstrate the effectiveness of the proposed method in unseen face forgery detection. Code is available at \url{https://github.com/OUC-VAS/ICSA}.

\end{abstract}
\section{Introduction}
\label{sec:Introduction}

Deepfake techniques have been misused to manipulate faces, raising serious societal concerns~\cite{ffpp2019,celebdf2020,survey2025}. As these techniques advance, new types of face forgeries continue to emerge. Differences in model architectures and training strategies lead to diverse forgery traces, posing a critical challenge for face forgery detection: models trained on existing datasets often struggle to detect unseen forgeries, limiting their effectiveness in practice.

Unsupervised domain adaptation offers a promising solution to address this challenge by transferring forgery knowledge learned from existing datasets (source domains) to unseen forgeries (target domains) using unlabeled target data~\cite{domainforensics2024,dpgnet2026,zhou2024,guo2025}. DomainForensics~\cite{domainforensics2024} is a representative method that adapts a single source model to a target domain. Seraj~\etal~\cite{icpr2025} explore adaptation from multiple source domains in a semi-supervised setting that uses a small number of labeled target samples. Despite their promise, they require access to source data during adaptation. In practice, however, data sharing restrictions, privacy requirements, and storage costs can make source data unavailable. This raises a natural question: \emph{Can we adapt multiple source models to detect unseen face forgeries without accessing their source data?}

\begin{figure}[t]
    \centering
    \includegraphics[
        page = 2,
        width=\columnwidth
    ]{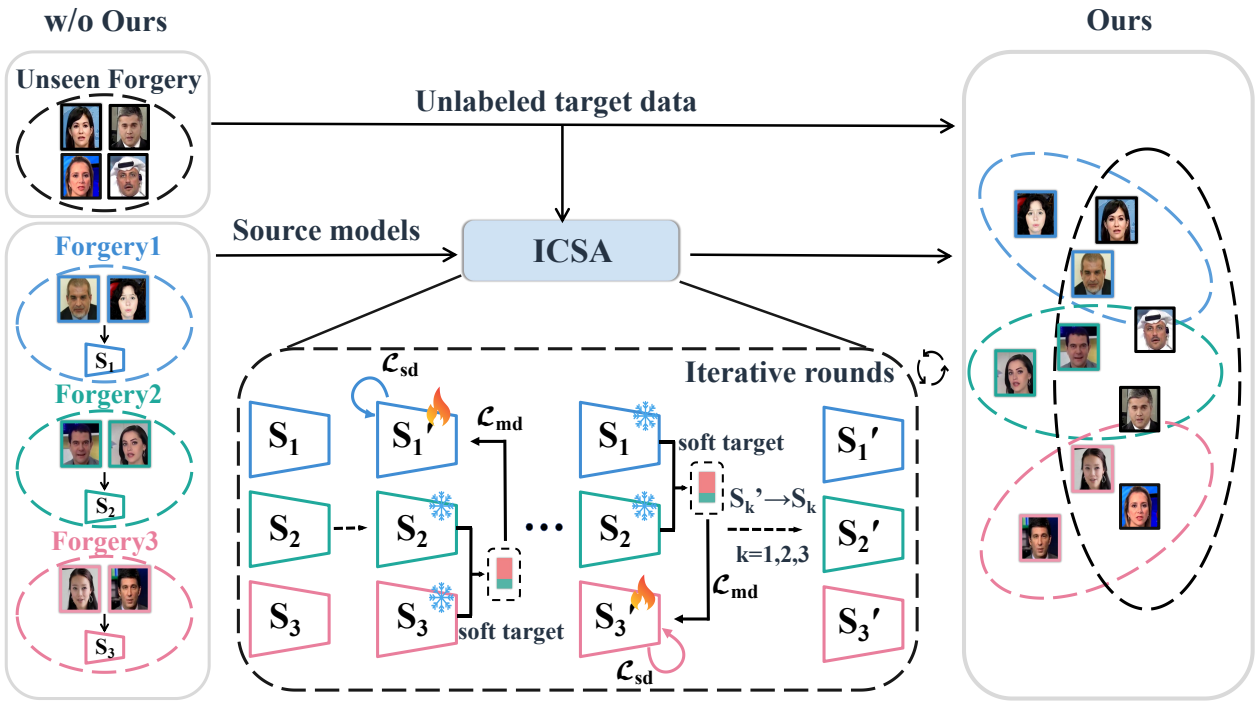}
    \caption{Overview of the proposed ICSA. Under the constraints of using only source models and unlabeled target data, ICSA narrows the gap between the unknown target distribution and the source distribution through alternation between collaborative distillation and iterative adaptation.}
    \label{fig:small_pipeline}
\end{figure}

Our goal aligns with a general vision task called multi-source-free domain adaptation (MSFDA). Previous studies have explored this problem through mechanisms such as model aggregation, pseudo-label construction, and transferability estimation~\cite{decision2021,caida2021,obbv2023,date2023,eaaf2024,biaten2024,zoo2025}. Nevertheless, these methods primarily address semantic classification, where adaptation preserves semantic knowledge across changes in appearance, \eg, from synthetic renderings to real images of cars. \textit{Face forgery detection presents a different challenge}: real and fake faces share similar semantic content, while their distinction relies on subtle forgery traces that vary across forgery methods. Because these traces are far less salient than semantic cues, existing adaptation methods struggle to effectively exploit source models for face forgery detection.

\begin{figure*}[!th]
    \centering
    \includegraphics[
        page=5,
        width=\textwidth
    ]{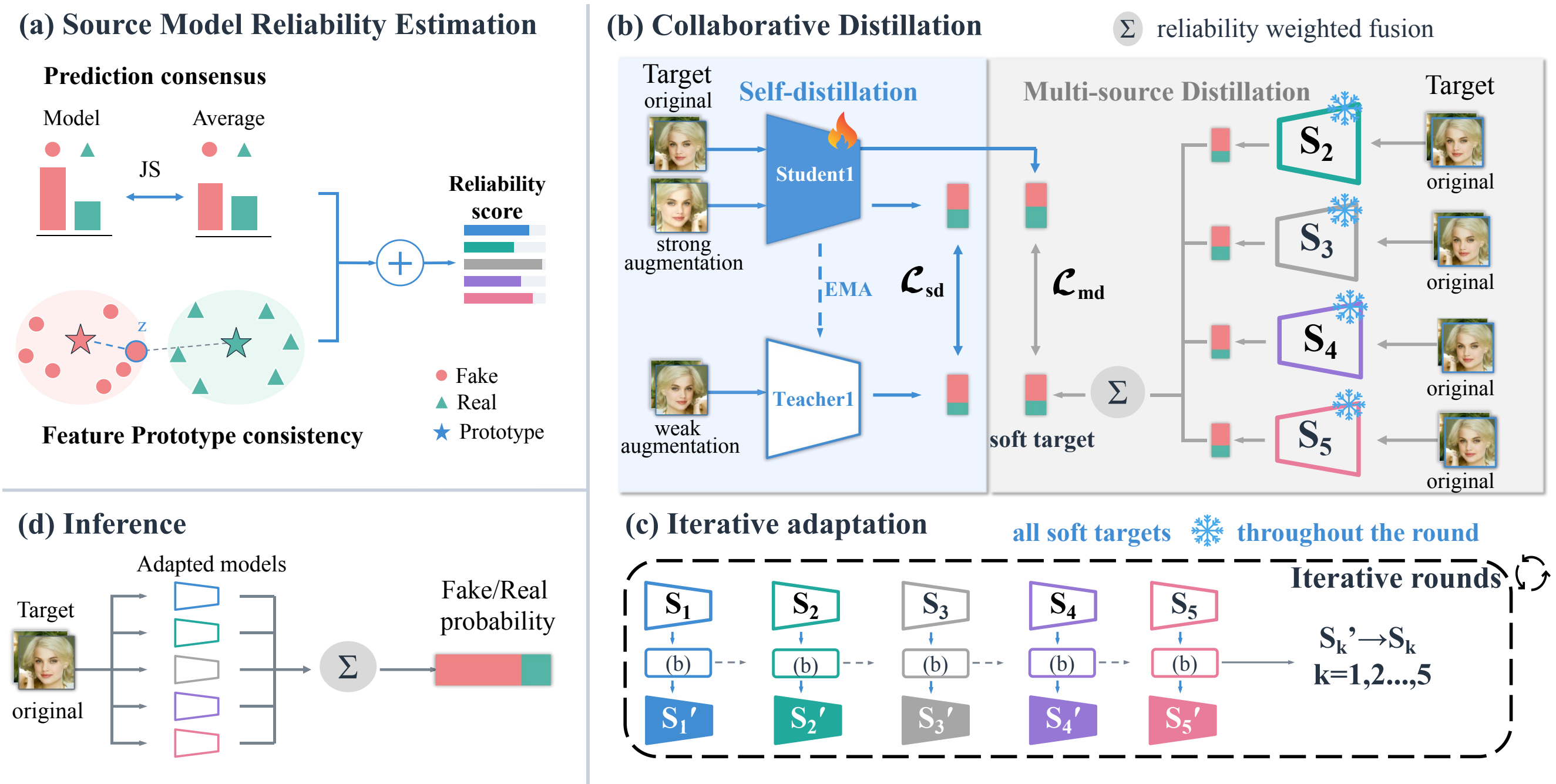}
    \vspace{-0.8cm}
    \caption{Pipeline of the proposed ICSA. We estimate the reliability of each source model on each target sample first in (a). Then we update the source models by collaborative distillation in (b) and perform iterative adaptation as shown in (c). In (d), we fuse the predictions of all adapted models to obtain final results.}
    \label{fig:framework}
\end{figure*}

To address this gap, we propose \textit{Iterative Collaborative Source-free Adaptation (ICSA)}, a multi-source adaptation framework for unseen face forgery detection without access to source data (Figure~\ref{fig:small_pipeline}). Our method consists of three key steps. First, \textit{Source Model Reliability Estimation} evaluates the reliability of each source model on target samples using prediction consistency and distances between target features and their corresponding prototypes. Second, \textit{Collaborative Distillation} jointly performs self-distillation and knowledge distillation guided by other source models. Third, \textit{Iterative Adaptation} alternates between reliability estimation and collaborative distillation to progressively refine the detectors. Together, this framework exploits the detection capabilities of source models and guides their adaptation by jointly considering model behavior and target data characteristics. At inference, ICSA fuses the predictions of all adapted models to obtain final results. Overview is shown in Figure~\ref{fig:framework}. 

To support systematic evaluation of this setting, we establish two protocols on the DF40 dataset~\cite{df402024} for within-category adaptation and cross-category adaptation. For within-category adaptation, we select six forgery methods from each of three forgery categories: Face Swapping (FS), Face Reenactment (FR), and Entire Face Synthesis (EFS). We designate each method in turn as the target domain and use five source models, each trained independently on corresponding data. This protocol yields 18 adaptation tasks. Cross-category adaptation draws source models from multiple forgery categories, making the protocol more challenging due to the substantial differences between categories. We compare our method with seven MSFDA baselines, surpassing the strongest by $4.94\%$ and $9.97\%$ under two protocols.

Our contributions are as follows: 
\begin{itemize}
\item To the best of our knowledge, we are the first to study source-free adaptation of multiple pretrained source detectors to unseen face forgeries using only unlabeled target data.
\item We propose ICSA, a dedicated framework that alternates between estimating source model reliability and distilling knowledge across models to improve detection of unseen forgeries.
\item We establish a DF40 evaluation protocol covering 18 adaptation tasks and compare ICSA with seven baselines, providing a benchmark for future work on source-free face forgery detection.
\end{itemize}

\section{Related Works}

\smallskip
\noindent\textbf{Face Forgery Detection.}
Face forgery involves using Deepfake techniques to forge faces, including face swapping, editing, and entire face synthesis. Because facial appearance conveys biometric identity, such manipulations can facilitate impersonation and other forms of misuse. To address these risks, numerous face forgery detectors have been proposed~\cite{li2018exposing,li2020face,shiohara2022detecting,tan2024rethinking,yan2025orthogonal,cui2025forensics}. These methods leverage various cues, such as biological signals~\cite{li2018ictu,ciftci2020fakecatcher,qi2020deeprhythm} and frequency-domain artifacts~\cite{qian2020thinking,luo2021generalizing,li2024freqblender,Kashiani_2025_CVPR}, or design specialized CNNs and Transformer architectures~\cite{afchar2018mesonet,dong2022protecting,zheng2021exploring}. Although these detectors perform well on public benchmarks, their performance often degrades on unseen forgeries.

To address this gap, several studies have explored unsupervised domain adaptation, which transfers knowledge from labeled source domains to an unlabeled target domain containing unseen forgeries~\cite{domainforensics2024,dpgnet2026,zhou2024,guo2025}. 
DomainForensics~\cite{domainforensics2024} combines adversarial alignment and self-distillation in a bidirectional adaptation process. DPGNet~\cite{dpgnet2026} uses text-guided cross-domain alignment and curriculum-based pseudo-label learning. Zhou~\etal~\cite{zhou2024} use adaptive clustering to generate fine-grained pseudo-labels that guide target feature learning. Guo~\etal~\cite{guo2025} combine domain distance optimization, class boundary separation, and adversarial learning to improve the discriminability and transferability of features across domains. Seraj~\etal~\cite{icpr2025} extend adaptation to multiple source domains in a semi-supervised setting that uses a small number of labeled target samples. 

Their common limitation is that they require access to labeled source data during adaptation, which may be impractical due to privacy or data-sharing restrictions. Other methods relax this requirement under different adaptation settings. 
FReTAL~\cite{fretal2021} transfers knowledge from a pretrained source model through distillation and feature matching but requires target labels. T\(^2\)A~\cite{t2a2025} uses uncertainty-aware learning to adapt a single source model online at test time. In this work, we stick to the unsupervised domain adaptation setting and study how to adapt multiple pretrained source detectors to unseen face forgeries using only unlabeled target data, without access to any source data. 

\smallskip
\noindent\textbf{Multi-Source-Free Domain Adaptation.}
Our goal falls under a task called multi-source-free domain adaptation (MSFDA), in which multiple pretrained source models are adapted using only unlabeled target data. Prior work on MSFDA has focused primarily on semantic image classification~\cite{decision2021,caida2021,obbv2023,eaaf2024,biaten2024}. DECISION~\cite{decision2021} combines information maximization with clustering-based pseudo-label generation, whereas CAiDA~\cite{caida2021} derives pseudo-labels from high-confidence anchors. OBBV~\cite{obbv2023} addresses label mismatch and feature misalignment through a hierarchical framework that combines pseudo-labeling based on cluster centers, Mixup augmentation, and joint feature alignment. DATE~\cite{date2023} weights source models according to their estimated discriminability and transferability. EAAF~\cite{eaaf2024} and Bi-ATEN~\cite{biaten2024} further account for how the relevance of each source model varies across target samples. SEA~\cite{zoo2025} relaxes the assumption that source models share a common architecture by adapting from a model zoo with diverse architectures.

Note that in semantic image classification, adaptation aims to preserve semantic information across domains. In contrast, in face forgery detection, authentic and forged faces share similar semantic content, and distinguishing them depends on subtle traces. Thus, strategies designed to preserve semantic class information likely overlook the cues needed to detect unseen forgeries. This motivates a dedicated framework for transferring forgery knowledge from multiple source detectors.

\section{Methodology}
\label{sec:Methodology}
We first formalize the problem setting and then present the three steps of ICSA (Figure~\ref{fig:framework}). First, we estimate the reliability of each source model on each target sample by considering two kinds of measures. Next, we update the source models sequentially through collaborative distillation using these estimates and target data. Finally, we repeat reliability estimation and collaborative distillation for a predefined number of rounds.
\subsection{Problem Setting}
\label{sec:problem_setting}
Denote $\{ S_k \}^{K}_{k=1}$ as $K$ trained face forgery detectors, \ie, source models. Each source model $S_k$ is trained on a labeled dataset $\mathcal{D}^{s}_k=\{(x_{k,i}^s,y_{k,i}^s)\}_{i=1}^{N^s_{k}}$ associated with a distinct forgery method, where $x_{k,i}^s$ is a face image and $y_{k,i}^s \in \{0,1\}$ is its label ($0$ for authentic and $1$ for forged). Denote $\mathcal D_t=\{x_i^t\}_{i=1}^{N^t}$ as unlabeled target data containing forgeries generated by a method unseen during source training. 
Our goal is to adapt source models $\{ S_k \}^{K}_{k=1}$ to detect unseen face forgeries using only $\mathcal D_t=\{x_i^t\}_{i=1}^{N^t}$, without access to any source data $\mathcal{D}_k^s$.

\subsection{Source Model Reliability Estimation}
\label{sec:reliability}
Source models trained on different types of face forgeries vary in their ability to detect unseen forgeries. Identifying which models are most reliable for each target sample is therefore critical for effective adaptation. To this end, we describe a strategy to assess source model reliability. 
Since forgery traces are particularly challenging, existing solutions designed for semantic classification~\cite{date2023,eaaf2024,biaten2024,zoo2025} may not be capable of learning these traces. To overcome this, we estimate reliability through two measures: Prediction Consensus Consistency and Feature Prototype Consistency. 

\smallskip
\noindent\textbf{Prediction Consensus Consistency.}
We assume that a source model is more reliable if its prediction is consistent with others.
Given target image $x_i^t$, we denote $h^t_{k,i}=S_k(x_i^t)$ as the logits of model $S_k$ and let $p_{k,i}=\mathrm{softmax}(h_{k,i})$ denote its predicted class probabilities. Since all predictions and features below are computed on target data, we omit the superscript $t$ from these quantities. This consistency can be defined as
\begin{equation}
\bar p_i =\frac{1}{K-1}\sum_{j\ne k}p_{j,i}, \;
\mathcal{A}_{k,i}=1 - \frac{1}{\alpha} \mathrm{JS}(p_{k,i},\bar p_i).
\label{eq:consensus}
\end{equation}
Here, $\mathrm{JS}$ denotes the Jensen--Shannon divergence~\cite{lin1991divergence}, defined as $\mathrm{JS}(p,q)=\tfrac{1}{2}\mathrm{KL}(p\|m)+\tfrac{1}{2}\mathrm{KL}(q\|m)$, where $m=(p+q)/2$. The normalization constant $\alpha$ is the maximum value of the divergence ($\log 2$ when using natural logarithms). Thus, $\mathcal{A}_{k,i}\in[0,1]$, with higher values indicating stronger agreement with the other models.

\smallskip
\noindent\textbf{Feature Prototype Consistency.}
Prediction consensus consistency alone can be misleading when several source models make the same incorrect prediction. To complement this measure,
we measure the behavior of the model itself by evaluating the consistency between the feature and the prototype of its predicted class. We assume a model is more reliable when its features are consistent with its corresponding prototype.
Let $z_{k,i}$ be the $\ell_2$-normalized feature from the last layer before the classifier of model $S_k$. For each class $c \in \{0,1\}$, we construct a probability-weighted mean of the target features and normalize it to obtain the class prototype: 
\begin{equation}
\tilde \mu^{(c)}_{k}=\frac{\sum_{i=1}^{N_t}p^{(c)}_{k,i}z_{k,i}}
{\sum_{i=1}^{N_t}p^{(c)}_{k,i}},\;
\mu^{(c)}_{k}=\frac{\tilde\mu^{(c)}_{k}}{\|\tilde\mu^{(c)}_{k}\|_2}.
\label{eq:score_prototypes}
\end{equation}
These are soft prototypes: every sample contributes to both classes according to its predicted probabilities. Denote $ c^{*}_{k,i}=\arg\max_c p^{(c)}_{k,i}$ as the class predicted by model $S_k$ for sample $i$. We define the feature prototype consistency score as
\begin{equation}
\mathcal{B}_{k,i}=\sigma\left(
\|z_{k,i}-\mu^{(1-c^{*}_{k,i})}_{k}\|_2^2
-\|z_{k,i}-\mu^{(c^{*}_{k,i})}_{k}\|_2^2\right),
\label{eq:alignment}
\end{equation}
where $\sigma$ is the sigmoid function. The score increases as the target feature becomes closer to the predicted-class prototype relative to the other prototype. We compute prototypes separately for each model and recompute them from the target data in every adaptation round. The final reliability score combines the two consistency measures:
\begin{equation}
    \mathcal{R}_{k,i} = \mathcal{A}_{k,i} + \mathcal{B}_{k,i}.
\end{equation}

\subsection{Collaborative Distillation}
\label{sec:collaborative distillation}
Using the reliability scores, we refine each source model through collaborative distillation, which combines multi-source distillation from other source models and self-distillation.

\smallskip
\noindent\textbf{Multi-Source Distillation.}
To take advantage of other source models, we introduce multi-source distillation to learn from their aggregated predictions. For each student model $S_k$, we treat the remaining $K-1$ source models as teachers and aggregate their predicted class probabilities into a soft target~\cite{dml2018}. For target sample $i$, we assign each teacher $S_j$ a weight based on its reliability score and obtain the teacher distribution as
\begin{equation}
\alpha_{j,i} =\frac{\exp(\mathcal{R}_{j,i}/\tau_w)}
{\sum_{l \neq k}\exp(\mathcal{R}_{l,i}/\tau_w)}, \; q_{k,i}=\sum_{j\ne k}\alpha_{j,i} p_{j,i}.
\label{eq:external_target}
\end{equation}
Here, $\alpha_{j,i}$ is the weight and $p_{j,i}$ denotes the predicted class probabilities of $S_j$ as in Eq.~\eqref{eq:consensus}, and $\tau_w>0$ controls how strongly differences in reliability affect the weights. The weights are normalized over the teachers, excluding the student $S_k$. We train $S_k$ to match this soft target by minimizing
\begin{equation}
\mathcal L_{\mathrm{md}} = \sum_{i}
\operatorname{KL}\left(q_{k,i} \,\middle\|\,
p_{k,i}\right).
\label{eq:mutual_loss}
\end{equation}
Note that $q_{k,i}$ are cached before each round and remain unchanged during student optimization.

\smallskip
\noindent\textbf{Self-Distillation.}
To complement knowledge transfer from other source models, we pair each student $S_k$ with a teacher $S'_k$ initialized with the student's parameters. The teacher tracks the student through an exponential moving average (EMA)~\cite{meanteacher2017}. Following DINO~\cite{dino2021}, we use two augmented views and center the teacher logits. For each target image $x_i$, the teacher receives a weakly augmented view and the student receives a strongly augmented view. Let $h_{k,i}^{\prime}$ and $h_{k,i}$ denote their respective logits. The corresponding class distributions are
\begin{equation}
    t_{k,i}
    = \operatorname{softmax}\!\left(
        \frac{h_{k,i}^{\prime}-b_k}{\tau_t}
    \right),
    \;
    s_{k,i}
    = \operatorname{softmax}\!\left(
        \frac{h_{k,i}}{\tau_s}
    \right),
\end{equation}
where $b_k$ is the running center of the teacher logits that is updated within a batch $B$ using 
$b_k\leftarrow m_c b_k+(1-m_c) \frac{1}{|B|}\sum_{i\in B}h'_{k,i}$, and $\tau_t$ and $\tau_s$ are the teacher and student temperatures, respectively. The self-distillation loss is the cross-entropy between the teacher and student distributions:
\begin{equation}
    \mathcal{L}_{\mathrm{sd}}
    = -\sum_i \sum_{c\in\{0,1\}}
        t_{k,i}^{(c)} \log s_{k,i}^{(c)}.
    \label{eq:self_loss}
\end{equation}
After each student optimization step, we update the teacher parameters as $S'_k\leftarrow m_tS'_k+(1-m_t)S_k$,
where $m_t$ is the teacher momentum. Each source model maintains its own teacher and logit center throughout adaptation.

The overall objective combines the two distillation losses:
\begin{equation}
    \mathcal{L}
    = \mathcal{L}_{\mathrm{sd}}
      + \lambda_{1}\mathcal{L}_{\mathrm{md}},
    \label{eq:total_loss}
\end{equation}
where $\lambda_{1}$ controls the contribution of multi-source distillation.

\subsection{Iterative Adaptation}
\label{sec:optimization_inference}
ICSA alternates between reliability estimation and collaborative distillation for $T$ rounds. At the start of round $r$, we evaluate the source models $\{S_k^{(r-1)}\}_{k=1}^{K}$ on all samples in $\mathcal{D}_t$ and compute their sample-level reliability scores $\mathcal{R}_{k,i}^{(r)}$ using \cref{eq:consensus,eq:score_prototypes,eq:alignment}.
We then train a student copy $\widetilde S_k$ of each source model in sequence, optimizing each copy for $n$ epochs through collaborative distillation. For student $k$ and target sample $i$, we use the reliability scores $\{\mathcal{R}_{j,i}^{(r)}\}_{j\ne k}$ to weight the predictions of the other $K-1$ source models and construct the teacher distribution $q_{k,i}^{(r)}$ according to \cref{eq:external_target}. The student learns to match this distribution through multi-source distillation while also performing self-distillation. The source models $\{S_k^{(r-1)}\}_{k=1}^{K}$ remain frozen throughout the round. We store each trained student copy separately and replace all source models simultaneously once every student has completed training, yielding $\{S_k^{(r)}\}_{k=1}^{K}$ for the next round.
After the final round, we recompute the reliability scores using the adapted models $\{S_j^{(T)}\}_{j=1}^{K}$. For each target sample $i$, we fuse the predictions of all $K$ models:
\begin{equation}
    \alpha_{j,i}
    = \frac{\exp(\mathcal{R}_{j,i}/\tau_w)}
           {\sum_{\ell}\exp(\mathcal{R}_{\ell,i}/\tau_w)},
    \quad
    q_i = \sum_{j}\alpha_{j,i}p_{j,i},
    \label{eq:inference}
\end{equation}
where $\mathcal{R}_{j,i}$ and $p_{j,i}$ denote the recomputed reliability score and predicted class probabilities of $S_j^{(T)}$, respectively. The final prediction is $\hat{y}_i=\arg\max_c q_i^{(c)}$. The full procedure is summarized in Algorithm~\ref{alg:icsa}.

\begin{algorithm}[t]
\small
\caption{\small Iterative Collaborative Source-Free Adaptation}
\label{alg:icsa}
\begin{algorithmic}[1]
\Require Source models $\{S_k^{(0)}\}_{k=1}^K$, unlabeled data
$\mathcal{D}_t$, rounds $T$, epochs $n$
\Ensure Adapted models and target predictions

\State Initialize each EMA teacher $S'_k \gets S_k^{(0)}$ and center $b_k \gets 0$
\For{$r=1,\ldots,T$}
    \State Freeze $\{S_k^{(r-1)}\}_{k=1}^K$ and estimate
           $\mathcal{R}_{k,i}^{(r)}$ on $\mathcal{D}_t$
    \State Cache teacher distributions $q_{k,i}^{(r)}$ using \cref{eq:external_target}
    \For{$k=1,\ldots,K$}
        \State Initialize $\widetilde S_k$ as a copy of $S_k^{(r-1)}$
        \State Train $\widetilde S_k$ for $n$ epochs using \cref{eq:total_loss}
        \Statex \hspace{\algorithmicindent}\hspace{\algorithmicindent}
               Update $S'_k$ and $b_k$ at each training step
    \EndFor
    \State $\{S_k^{(r)}\}_{k=1}^K \gets \{\widetilde S_k\}_{k=1}^K$
           \Comment{Replace all source models together}
\EndFor
\State Recompute reliability scores using $\{S_k^{(T)}\}_{k=1}^K$
\State Fuse predictions via \cref{eq:inference} and set $\hat y_i \gets \arg\max_c q_i^{(c)}$
\State \Return $\{S_k^{(T)}\}_{k=1}^K$, $\{\hat y_i\}_{i=1}^{N^t}$
\end{algorithmic}
\end{algorithm}

\section{Experiments}
\label{sec:Experiments}

\subsection{Experimental Settings}
\smallskip
\noindent\textbf{Datasets and Evaluation.}
We evaluate our method on DF40~\cite{df402024}, a benchmark comprising 40 forgery methods across multiple categories. We consider three categories, namely Face Swapping (FS), Face Reenactment (FR), and Entire Face Synthesis (EFS). For within-category adaptation, we select six methods per category that span diverse architectures and generation mechanisms to ensure variation in forgery cues and domain shifts (see Table~\ref{tab:selected_forgery_methods}). We designate each method in turn as the target domain and use five source models, each independently trained on the corresponding method. This protocol yields six adaptation tasks per category and 18 in total. We also evaluate under a cross-category protocol, which is more challenging because the source models are trained on forgery methods from categories that differ substantially in their forgery cues.

\begin{table}[!t]
  \centering
  \caption{The 18 selected DF40 forgery methods.}
  \vspace{-0.4cm}
  \label{tab:selected_forgery_methods}
    \small
  \setlength{\tabcolsep}{4pt}
  \resizebox{\linewidth}{!}{%
  \begin{tabular}{lll}
    \toprule
    {Face Swapping (FS)} &
    {Face Reenactment (FR)} &
    {Entire Face Synthesis (EFS)} \\
    \midrule
    BlendFace~\cite{shiohara2023blendface} & FOMM~\cite{siarohin2019fomm} & DiT-XL/2~\cite{peebles2023dit} \\
    FaceDancer~\cite{rosberg2023facedancer} & DaGAN~\cite{hong2022dagan} & RDDM~\cite{liu2024rddm}  \\
    SimSwap~\cite{chen2020simswap} & MCNet~\cite{hong2023mcnet} & DDPM~\cite{ho2020ddpm} \\
    e4s~\cite{liu2023e4s} & FS\_vid2vid~\cite{wang2019fs_vid2vid} & VQGAN~\cite{esser2021vqgan} \\
    InSwapper~\cite{inswapper} & SadTalker~\cite{zhang2023sadtalker} & StyleGAN3~\cite{karras2021stylegan3} \\
    MobileSwap~\cite{xu2022mobilefaceswap} & Wav2Lip~\cite{prajwal2020wav2lip} & StyleGAN-XL~\cite{sauer2022stylegan} \\
    \bottomrule
  \end{tabular}}
  \vspace{-0.4cm}
\end{table}

For a fair comparison, we use the preprocessed data provided by DF40 for all methods. All real videos originate from FaceForensics++ (FF++)~\cite{ffpp2019}, which provides 720 training videos and 140 videos for validation and testing. We reserve 140 of the original training videos for source-domain validation and testing, leaving 580 videos for source-domain training and retaining the original 140 videos for target-domain adaptation. The real videos in these three splits are mutually disjoint, so the source models do not encounter any of the real videos used for target-domain adaptation during source training. We report frame-level AUC on the entire target domain using the checkpoint from the final adaptation epoch.

\smallskip
\noindent\textbf{Baselines.}
To our knowledge, no existing method is specifically designed for face forgery detection. For a fair comparison, we adapt seven MSFDA methods to this task using their publicly released implementations and evaluate them under the same protocol. These methods include CAiDA (NeurIPS'21)~\cite{caida2021}, DECISION (CVPR'21)~\cite{decision2021}, OBBV (ICML'23)~\cite{obbv2023}, DATE (AAAI'23)~\cite{date2023}, EAAF (TPAMI'24)~\cite{eaaf2024}, Bi-ATEN (AAAI'24)~\cite{biaten2024}, and SEA (TPAMI'25)~\cite{zoo2025}.

\begin{table*}[!htb]
  \centering
  \caption{Performance under within-category adaptation. Among our method and the baselines, best results are in \textbf{bold}, and second-best results are \underline{underlined}. }
  \label{tab:main_table_all}
  \vspace{-0.3cm}
  \setlength{\tabcolsep}{3pt}
  \renewcommand{\arraystretch}{1.12}

  \resizebox{\textwidth}{!}{%
  \begin{tabular}{l|*{7}{c}|*{7}{c}|*{7}{c}|c}
    \toprule
     &
    \multicolumn{7}{|c}{{Face Swap (FS)}} &
    \multicolumn{7}{|c}{{Face Reenactment (FR)}} &
    \multicolumn{7}{|c|}{{Entire Face Synthesis (EFS)}} & \\
    \cmidrule(lr){2-8}\cmidrule(lr){9-15}\cmidrule(lr){16-22}

    Method & 
    \rotatebox{90}{$\rightarrow$MobileSwap} &
    \rotatebox{90}{$\rightarrow$e4s} &
    \rotatebox{90}{$\rightarrow$FaceDancer} &
    \rotatebox{90}{$\rightarrow$SimSwap} &
    \rotatebox{90}{$\rightarrow$InSwapper} &
    \rotatebox{90}{$\rightarrow$BlendFace} & Avg. &
    \rotatebox{90}{$\rightarrow$FOMM} &
    \rotatebox{90}{$\rightarrow$DaGAN} &
    \rotatebox{90}{$\rightarrow$FS\_vid2vid} &
    \rotatebox{90}{$\rightarrow$MCNet} &
    \rotatebox{90}{$\rightarrow$SadTalker} &
    \rotatebox{90}{$\rightarrow$Wav2Lip} & Avg. &
    \rotatebox{90}{$\rightarrow$DiT-XL/2} &
    \rotatebox{90}{$\rightarrow$RDDM} &
    \rotatebox{90}{$\rightarrow$StyleGAN-XL} &
    \rotatebox{90}{$\rightarrow$VQGAN} &
    \rotatebox{90}{$\rightarrow$StyleGAN3} &
    \rotatebox{90}{$\rightarrow$DDPM} & {Avg.} & {Avg.} \\
    \midrule

    Vanilla & 98.27
   & 81.53  & 74.82  & 98.50  & 84.85  & 97.93 & 89.32 & 100 & 99.93 & 100 & 99.95 & 95.96 & 67.60 & 93.91 & 46.67 & 96.02 & 57.46 & 93.16 & 57.06 & 93.56 & 73.99 & 85.74\\
    \midrule
    
    CAiDA~\cite{caida2021}  &
    91.93 & 73.83 & 53.05 & 78.19 & 56.60 & 81.45 & 72.51 &
    \textbf{100} & 97.61 & \textbf{100} & 98.80 & 64.65 & 51.09 & 85.36 &
    46.76 & \textbf{100} & 79.61 & 94.73 & 82.55 & 98.17 & 83.64 & 80.50 \\

    DECISION~\cite{decision2021}  &
    97.97 & \underline{86.59} & 60.19 & 91.98 & 72.40 & \underline{98.49} & 84.60 &
    99.71 & 99.31 & \textbf{100} & 99.70 & 84.32 & 50.55 & 88.93 &
    51.59 & \textbf{100} & 77.35 & \underline{97.81} & \textbf{100} & \textbf{99.49} & \underline{87.71} & \underline{87.08} \\

    OBBV~\cite{obbv2023}  &
    96.12 & 71.54 & 51.26 & \underline{97.47} & 68.70 & 98.47 & 80.59 &
    99.90 & 99.27 & \textbf{100} & 98.54 & 81.95 & 50.11 & 88.30 &
    50.39 & 99.64 & \underline{91.44} & 96.94 & 65.15 & 94.86 & 83.07 & 83.99 \\

    DATE~\cite{date2023}  &
    \textbf{98.58} & 85.82 & 58.84 & 86.58 & 71.74 & 98.28 & 83.31 &
    \underline{99.96} & \textbf{100} & \textbf{100} & \underline{99.79} & 77.50 & 52.91 & 88.36 &
    48.94 & \textbf{100} & 39.19 & 97.34 & 89.98 & 43.27 & 69.79 & 80.48 \\

    EAAF~\cite{eaaf2024}  &
    98.42 & 82.70 & 58.49 & 95.62 & \underline{81.36} & 98.06 & \underline{85.78} &
    \underline{99.96} & 99.50 & \underline{99.96} & 99.26 & \underline{88.37} & 53.97 & \underline{90.17} &
    45.90 & 99.91 & 42.75 & 96.27 & 59.74 & 97.95 & 73.75 & 83.23 \\

    Bi-ATEN~\cite{biaten2024}  &
    96.94 & 81.73 & 57.36 & 89.77 & 79.33 & 95.62 & 83.46 &
    \textbf{100} & 99.10 & \textbf{100} & 99.58 & 84.47 & \underline{55.57} & 89.79 &
    \underline{56.85} & \textbf{100} & 34.37 & 91.92 & 66.46 & 98.10 & 74.62 & 82.62 \\

    SEA~\cite{zoo2025}  &
    87.94 & 74.00 & \underline{69.96} & 79.06 & 74.50 & 81.15 & 77.77 &
    98.02 & 95.29 & 99.92 & 97.33 & 72.23 & -- & -- &
    -- & \underline{99.98} & 32.96 & 88.69 & 62.36 & 63.38 & -- & -- \\

    \midrule

    \textbf{Ours}  &
    \underline{98.50} & \textbf{91.80} & \textbf{72.13} & \textbf{98.86} & \textbf{84.80} & \textbf{98.98} & \textbf{90.85} &
    \textbf{100} & \underline{99.92} & \textbf{100} & \textbf{99.94} & \textbf{93.39} & \textbf{64.05} & \textbf{92.88} &
    \textbf{60.77} & \textbf{100} & \textbf{95.30} & \textbf{99.87} & \underline{99.33} & \underline{98.73} & \textbf{92.33} & \textbf{92.02} \\

    \bottomrule
  \end{tabular}%
  }
  \vspace{-0.4cm}
\end{table*}

\smallskip
\noindent\textbf{Source Models.}
We use Xception~\cite{xception2017} as the source model architecture in our main experiments because it is widely used for face forgery detection~\cite{ffpp2019,celebdf2020,qian2020thinking}. Following baselines~\cite{decision2021,caida2021,obbv2023,date2023,eaaf2024,biaten2024,zoo2025}, we insert a bottleneck layer between the feature extractor and the classification head to facilitate feature learning during adaptation. More architectures are studied in analysis experiments.

\smallskip
\noindent\textbf{Implementation Details.}
We sample 16 frames at fixed intervals from each video and resize them to $256\times256$. We train each source model independently for 10 epochs with a batch size of 32, using the same Albumentations augmentation pipeline with random cropping disabled.
By default, adaptation runs for $T=3$ rounds, with each student trained for $n=1$ epoch per round. We set the temperature for reliability weighting to $\tau_w=0.2$ and the weight for multi-source distillation to $\lambda_{\mathrm{1}}=0.1$. During adaptation, we use AdamW~\cite{adamw2019} with learning rates of $10^{-5}$ for the backbone and $10^{-4}$ for the bottleneck, a weight decay of $5\times10^{-5}$, and a batch size of 16. We update each teacher using an exponential moving average of its student’s parameters with momentum $m_t=0.996$. We set both teacher and student temperatures to $0.1$ and the momentum for updating the logit center to $m_c=0.9$. The teacher receives a weakly augmented view with random horizontal flipping. The student receives a more strongly augmented view that also applies blur, resizing, color perturbation, and JPEG compression. We run all experiments on a single NVIDIA GeForce RTX 3090.

\subsection{Results}

\label{sec:main_comparison}
\smallskip
\noindent\textbf{Within-Category Adaptation.}
Table~\ref{tab:main_table_all} reports frame-level AUC on FS, FR, and EFS. For each target, ``Vanilla'' averages the predictions of the five source models without adaptation. Among the baselines, only DECISION achieves a higher mean AUC than  ``Vanilla'', suggesting that existing adaptation methods is less effective for face forgery detection. Although our method also exhibits slight negative transfer on a few targets relative to ``Vanilla'', it ranks first or ties for first on 14 of the 18 target domains and second on the remaining four when compared to the seven baselines. It achieves the highest mean AUC in all three categories, with 90.85\% on FS, 92.88\% on FR, and 92.33\% on EFS. Across all 18 targets, our method achieves a mean AUC of 92.02\%, outperforming the strongest baseline, DECISION, by 4.94\%.
The gains are especially pronounced on targets in EFS such as DiT-XL/2, StyleGAN-XL, and StyleGAN3. A possible explanation is that synthesis traces are distributed across the image, allowing our method to combine complementary cues captured by different source models.
Note that SEA uses a Global Dispersity threshold to select source models. For Wav2Lip and DiT-XL/2, no source model meets this threshold, so SEA produces no result. We mark these entries with ``-'' in the table. 


\begin{table}[tb]
  \centering
  \caption{Performance under cross-category adaptation.}
  \label{tab:main_across_type}
  \vspace{-0.4cm}

  \begin{adjustbox}{max width=\columnwidth}
  \begin{tabular}{l|*{7}{c}}

    \toprule
    & \multicolumn{7}{|c}{Cross-Category} \\
    \cmidrule(lr){2-8}

    Method &
    \rotatebox{90}{$\rightarrow$InSwapper} &
    \rotatebox{90}{$\rightarrow$e4s} &
    \rotatebox{90}{$\rightarrow$StyleGAN-XL} &
    \rotatebox{90}{$\rightarrow$DiT-XL/2} &
    \rotatebox{90}{$\rightarrow$FOMM} &
    \rotatebox{90}{$\rightarrow$Wav2Lip} &
    \textbf{Avg.} \\

    \midrule

    Vanilla & 79.41
   & 76.25  & 76.92  & 64.01  & 93.44  & 65.86 & 75.98 \\
   
    \midrule
    
    CAiDA~\cite{caida2021} &
    58.13 & 64.94 & 80.14 & 53.83 &
    95.29 & 52.98 & 67.55 \\

    DECISION~\cite{decision2021} &
    \underline{65.85} & \underline{87.07} &
    84.98 & 53.16 & 96.27 & 52.17 &
    \underline{73.25} \\

    OBBV ~\cite{obbv2023} &
    51.04 & 66.77 & 57.52 &
    \underline{60.77} & 75.46 & 51.53 & 60.52 \\

    DATE ~\cite{date2023} &
    62.24 & 82.77 & \underline{97.42} &
    49.29 & 96.41 & 48.90 & 72.84 \\

    EAAF ~\cite{eaaf2024} &
    63.61 & 65.88 & 88.63 & 52.95 &
    96.27 & 55.61 & 70.49 \\

    Bi-ATEN ~\cite{biaten2024} &
    58.91 & 71.42 & \textbf{98.46} &
    54.80 & \textbf{98.33} & 55.54 & 72.91 \\

    SEA ~\cite{zoo2025} &
    58.67 & 52.93 & 89.99 & 58.59 &
    94.22 & \underline{61.44} & 69.31 \\

    \midrule

    Ours &
    \textbf{79.87} & \textbf{93.74} &
    94.77 & \textbf{63.53} &
    \underline{98.20} & \textbf{69.21} &
    \textbf{83.22} \\

    \bottomrule

  \end{tabular}
  \end{adjustbox}

  \vspace{-0.3cm}
\end{table}

\smallskip
\noindent\textbf{Cross-Category Adaptation.}
To evaluate adaptation under larger domain shifts, we extend our within-category protocol to cross-category adaptation. Each target domain is paired with five source models. One is trained on a forgery method from the same category as the target, and the other four are trained on methods from the other two categories. 
As shown in Table~\ref{tab:main_across_type}, our method achieves a mean AUC of 83.22\%, outperforming DECISION by 9.97\%. It achieves the highest AUC on four of the six targets, with gains over the best baseline on InSwapper, e4s, and Wav2Lip of 14.02\%, 6.67\%, and 7.77\%, respectively. 
Our method outperforms ``Vanilla'' on five of the six targets and is the only one to surpass ``Vanilla'' in mean AUC, demonstrating a clear advantage over existing baselines under larger domain shifts. 
Interestingly, we observe that the AUC scores on Wav2Lip, DiT-XL/2, and e4s increase by 5.16\%, 2.76\%, and 1.94\%, respectively, compared to the within-category protocol. 
This suggests that the same forgery category does not necessarily imply similar forgery artifacts. The weighted fusion of diverse source models may provide more generalizable cues from diverse source models.

\subsection{Analysis}
We conduct all analyses on the 18 targets under the within-category protocol. Due to space constraints, we report results for the most challenging category, Entire Face Synthesis (EFS), in the main paper. Full results are provided in the \textit{supplementary}.

\smallskip
\noindent\textbf{Analysis of Distillation.}
\label{sec:loss_ablation}
Table~\ref{tab:ablation} evaluates the effects of self-distillation and multi-source distillation. Uniformly averaging the predictions of the source models before adaptation yields a mean AUC of 73.99\%. Self-distillation alone ($\mathcal{L}_{\mathrm{sd}}$) improves AUC on all six targets and raises the mean to 90.36\%. Adding multi-source distillation ($\mathcal{L}_{\mathrm{md}}$) yields the highest mean AUC of 92.33\%, a further gain of 1.97\%.
These results suggest that supervision from other source models can complement self-distillation, with benefits that vary across target domains.
In Table~\ref{tab:kd_weight}, we analyze the sensitivity of the distillation weight $\lambda_1$. It can be seen that $\lambda_1=0.1$ gives the highest mean of 92.33\%; increasing it to 1.0 reduces the mean to 87.44\%. Stronger external supervision therefore does not consistently improve adaptation. 
\begin{table}[!htb]
  \vspace{-0.4cm}
  \centering
  \caption{Effects of $\mathcal{L}_{\mathrm{sd}}$ and $\mathcal{L}_{\mathrm{md}}$.}
  \label{tab:ablation}
  \vspace{-0.4cm}
  \small
  \begin{adjustbox}{max width=\columnwidth}
  \begin{tabular}{cc|ccccccc}
  
    \toprule

    \multicolumn{2}{c}{} &
    \multicolumn{7}{|c}{Entire Face Synthesis (EFS)}\\
    
    \cmidrule(lr){3-9}

   $\mathcal{L}_{\mathrm{sd}}$&  $\mathcal{L}_{\mathrm{md}}$& 
    \rotatebox{90}{$\rightarrow$DiT-XL/2} &
    \rotatebox{90}{$\rightarrow$RDDM} &
    \rotatebox{90}{$\rightarrow$StyleGAN-XL} &
    \rotatebox{90}{$\rightarrow$VQGAN} &
    \rotatebox{90}{$\rightarrow$StyleGAN3} &
    \rotatebox{90}{$\rightarrow$DDPM} & Avg. \\
   
    \midrule

    & & 46.67 & 96.02 & 57.46 & 93.16 & 57.06 & 93.56 & 73.99 \\

    $\checkmark$ & & 61.31 & 100 & 85.39 & 99.83 & 95.84 & 99.78 & 90.36 \\

    $\checkmark$ & $\checkmark$ & 60.77 & 100 & 95.30 & 99.87 & 99.33 & 98.73 & 92.33 \\





    
    \bottomrule
    
  \end{tabular}%
  \end{adjustbox}
  \vspace{-0.4cm}
\end{table}
\begin{table}[!htb]
  \centering
  \caption{Sensitivity to distillation weight $\lambda_1$.}
  \label{tab:kd_weight}
  \vspace{-0.4cm}
  \begin{adjustbox}{max width=\columnwidth}
  \begin{tabular}{c|ccccccc}
  
    \toprule
     & 
    \multicolumn{6}{c}{Entire Face Synthesis (EFS)} \\
    \cmidrule(lr){2-8}
    
     $\lambda_1$& 
    \rotatebox{90}{$\rightarrow$DiT-XL/2} & 
    \rotatebox{90}{$\rightarrow$RDDM} & 
    \rotatebox{90}{$\rightarrow$StyleGAN-XL} & 
    \rotatebox{90}{$\rightarrow$VQGAN} & 
    \rotatebox{90}{$\rightarrow$StyleGAN3} & 
    \rotatebox{90}{$\rightarrow$DDPM} & Avg.\\
  
    \midrule

    0.05 & 61.53 & 100 & 93.91 & 99.85 & 99.30 & 98.97 & 92.26 \\

    0.10 (Ours) & 60.77 & 100 & 95.30 & 99.87 & 99.33 & 98.73 & 92.33 \\

    0.30 & 59.28 & 99.99 & 83.02 & 99.88 & 98.84 & 98.83 & 89.97 \\

    0.50 & 59.44 & 100 & 86.92 & 99.90 & 96.81 & 98.98 & 90.34 \\

    1 & 58.55 & 100 & 73.51 & 99.94 & 93.24 & 99.42 & 87.44 \\

    \bottomrule

  \end{tabular}
  \end{adjustbox}
\end{table} 

\smallskip
\noindent\textbf{Self-Distillation Design Choices.}
\label{sec:self-distillation measures}
Table~\ref{tab:self_mechanism} evaluates teacher centering (TC), weak teacher and strong student augmentations (WS), frozen batch normalization statistics (FBN), and symmetric temperatures (ST). With all four enabled, the mean AUC reaches 92.33\%. Removing TC causes the largest drop, to 76.22\%, with AUC falling from 95.30\% to 42.96\% on StyleGAN-XL and from 99.33\% to 71.37\% on StyleGAN3. Using weak augmentations for both networks or updating batch normalization statistics reduces the mean AUC to 86.49\% and 85.41\%, respectively. Lowering the teacher temperature from 0.1 to 0.04 while keeping the student temperature at 0.1, following DINO~\cite{dino2021}, reduces the mean AUC to 88.19\%. These results suggest that centering the teacher's predictions and avoiding excessive sharpening can help stabilize self-distillation on unseen forgeries.

\begin{table}[!htb]
  \centering
  \caption{Analysis of self-distillation design choices.
}
  \vspace{-0.4cm}
  \label{tab:self_mechanism}
  \begingroup
  
  \setlength{\tabcolsep}{3pt} 

  \begin{adjustbox}{max width=\columnwidth}

  \begin{tabular}{cccc|ccccccc}
  
    \toprule
    
     & & & &
     \multicolumn{7}{c}{Entire Face Synthesis (EFS)} \\
     
    \cmidrule(lr){5-11}
    
     TC& WS & FBN & ST& 
    \rotatebox{90}{$\rightarrow$DiT-XL/2} &
    \rotatebox{90}{$\rightarrow$RDDM} &
    \rotatebox{90}{$\rightarrow$StyleGAN-XL} &
    \rotatebox{90}{$\rightarrow$VQGAN} &
    \rotatebox{90}{$\rightarrow$StyleGAN3} &
    \rotatebox{90}{$\rightarrow$DDPM} & Avg. \\
  
    \midrule

    & $\checkmark$ & $\checkmark$ & $\checkmark$ & 43.98 & 99.87 & 42.96 & 99.40 & 71.37 & 99.75 & 76.22 \\

    $\checkmark$ & & $\checkmark$ & $\checkmark$ & 62.46 & 99.98 & 79.31 & 97.74 & 83.15 & 96.31 & 86.49 \\

    $\checkmark$ & $\checkmark$ & & $\checkmark$ & 58.78 & 100 & 78.40 & 99.92 & 77.28 & 98.09 & 85.41 \\

    $\checkmark$ & $\checkmark$ & $\checkmark$ & & 59.34 & 100 & 83.19 & 99.78 & 88.16 & 98.64 & 88.19 \\

    $\checkmark$ & $\checkmark$ & $\checkmark$ & $\checkmark$ & 60.77 & 100 & 95.30 & 99.87 & 99.33 & 98.73 & 92.33 \\

    \bottomrule
    
  \end{tabular}%
  \end{adjustbox}
  \endgroup
\end{table}

\smallskip
\noindent\textbf{Iterative Adaptation Strategies.}
Table~\ref{tab:refresh_strategy} compares five strategies for adapting the source models and updating their supervision. ``Joint'' optimizes all five models simultaneously with a single loss, with aggregation weights and fused soft targets refreshed after every batch. Soft target is explained in Sec.~\ref{sec:collaborative distillation}. ``Fixed weights'' retains the initial aggregation weights throughout adaptation. ``Per-model'' and ``Per-batch'' refresh the weights and soft targets after each model update and each batch, respectively. ``Per-round'' (ours) refreshes them after all five models complete the round. 
It can be seen that our ``Per-round'' achieves the highest mean AUC among the five strategies. Fixed weights may become less reliable as source models adapt, while joint optimization and frequent supervision updates may propagate transient prediction errors across models. Updating once per round gives each model a stable supervisory signal within the round and allows the models to exchange updated predictions between rounds.

\begin{table}[!htb]
  \centering
  \caption{
    Analysis of iterative adaptation strategies. Time denotes the cumulative training time across all 18 domains. 
  }
  \label{tab:refresh_strategy}
  \vspace{-0.4cm}
  \setlength{\tabcolsep}{2.2pt}

  \begin{adjustbox}{max width=\columnwidth}
  \begin{tabular}{l|cccccccc}
    \toprule
     &
    \multicolumn{6}{c}{Entire Face Synthesis (EFS)}\\
    \cmidrule(lr){2-8}
    
    Strategy& 
    \rotatebox{90}{$\rightarrow$DiT-XL/2} &
    \rotatebox{90}{$\rightarrow$RDDM} &
    \rotatebox{90}{$\rightarrow$StyleGAN-XL} &
    \rotatebox{90}{$\rightarrow$VQGAN} &
    \rotatebox{90}{$\rightarrow$StyleGAN3} &
    \rotatebox{90}{$\rightarrow$DDPM} & Avg.& 
    \mbox{Time (h) $\downarrow$}
    \\
    \midrule

    Joint & 57.63 & 100 & 84.87 & 99.88 & 97.04 & 99.31 & 89.79 & 8.42 \\

    Fixed weights & 60.35 & 99.01 & 94.73 & 99.60 & 98.10 & 95.39 & 91.20 & 3.90 \\

    Per-model & 59.29 & 100 & 91.91 & 99.78 & 96.17 & 99.15 & 91.05 & 4.97 \\

    Per-batch & 60.28 & 100 & 93.97 & 99.86 & 95.50 & 99.43 & 91.51 & 9.82 \\

    Per-round  & 60.77 & 100 & 95.30 & 99.87 & 99.33 & 98.73 & 92.33 & 4.72 \\
    \bottomrule
  \end{tabular}
  \end{adjustbox}
\end{table} 

\smallskip
\noindent\textbf{Reliability Estimation Measures.}
Table~\ref{tab:reliability} evaluates prediction consensus and feature prototype consistency. With both measures disabled, all models receive equal weights during adaptation and inference. Prediction consensus alone yields the highest AUC on DiT-XL/2, while feature prototype consistency alone performs best on StyleGAN3. Combining the two yields the highest mean AUC, suggesting that they provide complementary reliability cues at low computational cost.

\begin{table}[!htb]
  \centering
  \caption{Analysis of reliability estimation measures.}
  \label{tab:reliability}
  \vspace{-0.4cm}
  \begin{adjustbox}{max width=\columnwidth}
  \begin{tabular}{l|ccccccc}
  
    \toprule
    
     & \multicolumn{6}{c}{Entire Face Synthesis (EFS)}\\
    \cmidrule(lr){2-8}
    
     measures& 
    \rotatebox{90}{$\rightarrow$DiT-XL/2} &
    \rotatebox{90}{$\rightarrow$RDDM} &
    \rotatebox{90}{$\rightarrow$StyleGAN-XL} &
    \rotatebox{90}{$\rightarrow$VQGAN} &
    \rotatebox{90}{$\rightarrow$StyleGAN3} &
    \rotatebox{90}{$\rightarrow$DDPM} & Avg.\\
  
    \midrule

    Equal weights & 60.61 & 100 & 92.17 & 99.85 & 97.24 & 99.49 & 91.56 \\

    $+\mathcal{A}$ & 61.00 & 100 & 92.12 & 99.83 & 96.48 & 99.42 & 91.48 \\

    $+\mathcal{B}$ & 60.58 & 99.99 & 93.57 & 99.84 & 99.54 & 99.39 & 92.15 \\

    $+\mathcal{A}+\mathcal{B}$ & 60.77 & 100 & 95.30 & 99.87 & 99.33 & 98.73 & 92.33 \\

    \bottomrule
    
  \end{tabular}
  \end{adjustbox}
\end{table}


\smallskip
\noindent\textbf{Adaptation Using Various Backbones.}
\label{sec:backbone_transfer}
Table~\ref{tab:backbone} evaluates our adaptation method on Xception~\cite{xception2017}, ResNet-50~\cite{resnet2016}, EfficientNet-B4~\cite{efficientnet2019}, Swin-Tiny~\cite{swin2021}, and DeiT-Small/16~\cite{deit2021}. Our method improves mean AUC across all five backbones, with Xception showing the largest gain and EfficientNet-B4 achieving the highest mean AUC after adaptation. Under comparable parameter budgets, the evaluated CNN backbones achieve higher cross-domain performance than the Transformer backbones under our protocol.

\begin{table}[!htb]
  \centering
  \small
  \caption{Analysis of adaptation using various backbones.}
  \label{tab:backbone}
  \vspace{-0.4cm}
  \begin{adjustbox}{max width=\columnwidth}
    \begin{tabular}{lcc}
      \toprule
      Backbone        & EFS Avg. & Gain \\
      \midrule
      Xception~\cite{xception2017}        & 73.99 $\to$ 92.33    & +18.34 \\
      ResNet-50~\cite{resnet2016}      & 73.36 $\to$ 82.43    & +9.07  \\
      EfficientNet-B4~\cite{efficientnet2019} & 89.36 $\to$ 94.81    & +5.45  \\
      Swin-Tiny ~\cite{swin2021}      & 65.03 $\to$ 72.86    & +7.83  \\
      DeiT-Small/16 ~\cite{deit2021}  & 60.12 $\to$ 72.56    & +12.44 \\
      \bottomrule
    \end{tabular}
  \end{adjustbox}
\end{table}

  
  
    
  




    



\begin{figure}[!ht]
    \centering
    \begin{subfigure}[b]{0.48\columnwidth}
        \centering
        \includegraphics[width=\linewidth]{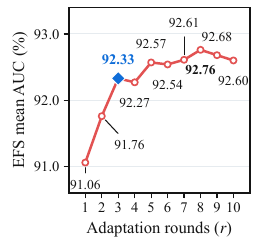}
        \label{fig:rounds}
    \end{subfigure}
    \hfill 
    \begin{subfigure}[b]{0.48\columnwidth}
        \centering
        \includegraphics[width=\linewidth]{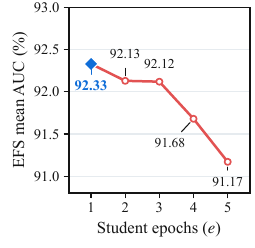}
        \label{fig:epochs}
    \end{subfigure}
    \vspace{-0.8cm}
    \caption{Number of rounds and student epochs.}
    \vspace{-0.cm}
    \label{fig:rounds-and-epochs}
\end{figure}

\smallskip
\noindent\textbf{Number of Rounds and Student Epochs.}
\label{sec:sensitivity}
In Figure~\ref{fig:rounds-and-epochs} (left), mean AUC rises from
91.06\% at one round to 92.33\% at the default three rounds, peaks
at 92.76\% at eight rounds, and dips to 92.60\% at ten rounds.
Mean AUC peaks at 92.33\% with one student epoch per round and
declines with additional epochs (right), possibly because less
frequent soft target updates weaken corrective guidance.

\begin{figure}[!ht]
    \centering
    \includegraphics[width=\columnwidth]{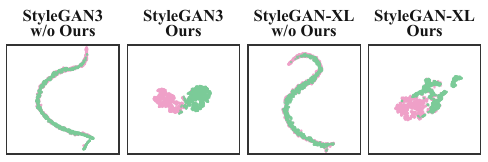}
    \vspace{-0.8cm}
    \caption{The t-SNE visualization~\cite{tsne2008} of StyleGAN-XL and StyleGAN3 based on their target data. \textcolor{dotpink}{\ensuremath{\bullet}} real, \textcolor{dotgreen}{\ensuremath{\bullet}} fake.}
    \label{fig:tsne}
\end{figure}

\begin{figure}[!ht]
    \centering
    \includegraphics[width=\columnwidth]{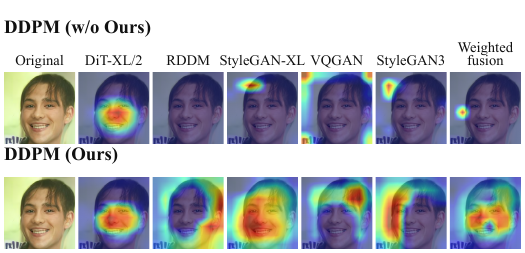}
    \vspace{-0.8cm}
    \caption{Grad-CAM~\cite{selvaraju2017grad} visualization with and without our adaptation method.}
    \label{fig:cam}
\end{figure}

\smallskip
\noindent\textbf{GradCam and t-SNE Analysis.}
Figure~\ref{fig:tsne} compares the feature distributions before and after adaptation on the StyleGAN3 and StyleGAN-XL target domains. For each domain, we use the same 2,000 images before and after adaptation and visualize the bottleneck features aggregated using the reliability weights. The real and fake clusters
overlap less after adaptation, suggesting improved class separability.
On DDPM, the fake-class Grad-CAM maps in Figure~\ref{fig:cam} show that
the fused response shifts from a small background region to the face
and its surroundings after adaptation, combining responses from
source models that attend to different regions.


\smallskip
\noindent\textbf{Limitations.}
Since face forgery traces can vary substantially across methods, certain source models may have a large gap to the target domain. Including these models likely leads to negative transfer. Future work will explore criteria to exclude source models with unadapted gaps.  
\section{Conclusion}
\label{sec:Conclusion}
We propose Iterative Collaborative Source-Free Adaptation (ICSA), a framework that adapts multiple source models to detect unseen face forgeries without access to source data. ICSA has three components: 1) Source Model Reliability Estimation assesses the reliability of each model on individual target samples. 2) Collaborative Distillation adapts each model by combining self-distillation and multi-source distillation. 3) Iterative Adaptation progressively refines the models by alternating reliability estimates and collaborative distillation. Experimental results demonstrate the effectiveness of ICSA in detecting unseen face forgeries.

{
    \small
    \putbib[main]
}

\clearpage
\end{bibunit}

\begin{bibunit}[ieeenat_fullname]

\appendix

\noindent{\Large\bfseries Supplementary Material\par}
\begin{table*}[tb]
  \centering
  \caption{The 18 selected DF40 forgery methods, grouped by forgery category and subtype. }
  \label{tab:selected_methods}
    \vspace{-0.3cm}
    \footnotesize
  \begin{adjustbox}{max width=\textwidth}
  \begin{tabular}{ccccccccc}
    \toprule
    \multicolumn{3}{c}{Face swapping (FS)} &
    \multicolumn{3}{c}{Face reenactment (FR)} &
    \multicolumn{3}{c}{Entire face synthesis (EFS)} \\
    \cmidrule(lr){1-3}\cmidrule(lr){4-6}\cmidrule(lr){7-9}
    Method & Subtype & Vanilla & Method & Subtype & Vanilla & Method & Subtype & Vanilla \\
    \midrule
    BlendFace & Disentanglement & 97.93 & FOMM & Image-driven & 100 & DiT-XL/2 & Diffusion-based & 46.67 \\
    FaceDancer & Disentanglement & 74.82 & DaGAN & Image-driven & 99.93 & RDDM & Diffusion-based & 96.02 \\
    SimSwap & Disentanglement & 98.50 & MCNet & Image-driven & 99.95 & DDPM & Diffusion-based & 93.56 \\
    e4s & Disentanglement & 81.53 & FS\_vid2vid & Landmark-driven & 100 & VQGAN & GAN-based & 93.16 \\
    InSwapper & Used in Roop & 84.85 & SadTalker & Audio-driven & 95.96 & StyleGAN3 & GAN-based & 57.06 \\
    MobileSwap & Lightweight & 98.27 & Wav2Lip & Audio-driven & 67.60 & StyleGAN-XL & GAN-based & 57.46 \\
    \bottomrule
  \end{tabular}
  \end{adjustbox}
\end{table*}

\section{Additional Explanations}
\subsection{Compared Methods}
We review several domain adaptation methods for face forgery detection in the related works ([\blue{32}, \blue{60}, \blue{64}, \blue{11}] in the main paper). These methods require either target labels or source data, making them incompatible with our source-free unsupervised domain adaptation setting. We attempted to modify their implementations for our setting, but their core mechanisms depend on these unavailable inputs and could not function without them. We therefore exclude these methods from our comparisons and instead adapt existing MSFDA methods to face forgery detection.


\subsection{The Selection of Forgery Methods}
We select these forgery methods based on three considerations.
\begin{enumerate}
    \item \textbf{Coverage of generation mechanisms.}
    The selected methods span face swapping, face reenactment, and entire face synthesis, covering diverse generation workflows (\eg, disentanglement-based, keypoint-driven, and audio-driven) and architectures (\eg, GAN-based and diffusion-based), as summarized in Table~\ref{tab:selected_methods}. They produce distinct forgery traces, ranging from localized artifacts to traces distributed throughout the image, introducing diverse domain shifts.

    \item \textbf{Variation in adaptation difficulty.} As described in the main paper, ``Vanilla'' averages the predictions of the five source models without adaptation. Table~\ref{tab:selected_methods} shows various adaptation difficulties across the targets. Vanilla performance is already close to saturation on visually driven reenactment and several face swapping targets, but remains considerably lower on Wav2Lip, StyleGAN-XL, StyleGAN3, and DiT-XL/2. Even within the same subtype, differences in workflow and architectures can result in low cross-method AUC. As shown in Tables~\ref{tab:supp_fs},~\ref{tab:supp_FR} and~\ref{tab:supp_efs}, the within-category Vanilla AUCs for disentanglement-based face swapping, audio-driven reenactment, diffusion-based synthesis, and GAN-based synthesis are only 74.81\%, 71.15\%, 69.53\%, and 62.95\%, respectively. Including both easier and harder targets allows us to assess improvements on difficult shifts together with performance retention on more transferable domains.

    \item \textbf{Task suitability.} Each target domain is associated with five source models with varying transferability, where highly transferable source models often coexist with models performing near random guessing or even exhibiting reversed predictions. This diversity supports both within-category and cross-category evaluation, allowing us to assess how effectively adaptation methods exploit useful source knowledge
\end{enumerate}

\begin{table*}[tp]
  \centering
  \caption{Within-category adaptation on Face Swap (FS). Cells show scores before and after adaptation.}
  \label{tab:supp_fs}
  \vspace{-0.3cm}
  \resizebox{\textwidth}{!}{%
    \begin{tabular}{l|*{6}{c}|c}
      \toprule
       & \multicolumn{6}{|c|}{Source models} & \\
      \cmidrule(lr){2-7}
      Target & MobileSwap & e4s & FaceDancer & SimSwap & InSwapper & BlendFace & Vanilla $\to$ Ours \\
      \midrule
      MobileSwap & --- & 68.96 $\to$ 73.75 & 79.82 $\to$ 84.13 & 90.93 $\to$ 96.57 & 97.07 $\to$ 99.16 & 96.62 $\to$ 97.62 & 98.27 $\to$ 98.50 \\
      e4s & 75.93 $\to$ 86.62 & --- & 79.35 $\to$ 84.99 & 62.09 $\to$ 76.70 & 70.69 $\to$ 84.08 & 80.69 $\to$ 90.98 & 81.53 $\to$ 91.80 \\
      FaceDancer & 63.37 $\to$ 61.37 & 71.06 $\to$ 76.79 & --- & 60.92 $\to$ 64.39 & 71.77 $\to$ 74.39 & 60.18 $\to$ 58.71 & 74.82 $\to$ 72.13 \\
      SimSwap & 89.13 $\to$ 94.70 & 67.27 $\to$ 75.20 & 77.41 $\to$ 83.47 & --- & 98.30 $\to$ 99.45 & 95.04 $\to$ 97.83 & 98.50 $\to$ 98.86 \\
      InSwapper & 69.49 $\to$ 68.16 & 60.92 $\to$ 62.85 & 79.98 $\to$ 83.36 & 75.21 $\to$ 85.21 & --- & 81.73 $\to$ 85.06 & 84.85 $\to$ 84.80 \\
      BlendFace & 92.22 $\to$ 97.61 & 72.18 $\to$ 86.64 & 75.92 $\to$ 82.66 & 95.62 $\to$ 98.85 & 95.81 $\to$ 98.99 & --- & 97.93 $\to$ 98.98 \\
      \bottomrule
    \end{tabular}%
  }
\end{table*}

\begin{table*}[tp]
  \centering
  \caption{Within-category adaptation on Face Reenactment (FR). Cells show scores before and after adaptation.}
  \label{tab:supp_FR}
  \vspace{-0.3cm}

  \resizebox{\textwidth}{!}{%
    \begin{tabular}{l|*{6}{c}|c}
      \toprule
       & \multicolumn{6}{|c|}{Source models} & \\
      \cmidrule(lr){2-7}
      Target & FOMM & DaGAN & FS\_vid2vid & MCNet & SadTalker & Wav2Lip & Vanilla $\to$ Ours \\
      \midrule
      FOMM & --- & 99.94 $\to$ 100 & 99.21 $\to$ 99.86 & 99.92 $\to$ 100 & 99.89 $\to$ 100 & 61.41 $\to$ 77.60 & 100 $\to$ 100 \\
      DaGAN & 97.61 $\to$ 99.15 & --- & 97.36 $\to$ 98.51 & 99.99 $\to$ 99.98 & 98.89 $\to$ 99.89 & 75.17 $\to$ 87.16 & 99.93 $\to$ 99.92 \\
      FS\_vid2vid & 99.99 $\to$ 100 & 99.96 $\to$ 100 & --- & 99.99 $\to$ 100 & 99.76 $\to$ 100 & 71.69 $\to$ 90.87 & 100 $\to$ 100 \\
      MCNet & 98.12 $\to$ 99.30 & 99.97 $\to$ 99.99 & 97.50 $\to$ 98.66 & --- & 99.18 $\to$ 99.72 & 79.56 $\to$ 90.69 & 99.95 $\to$ 99.94 \\
      SadTalker & 80.60 $\to$ 83.57 & 97.51 $\to$ 95.87 & 68.47 $\to$ 74.47 & 96.52 $\to$ 94.80 & --- & 69.42 $\to$ 76.07 & 95.96 $\to$ 93.39 \\
      Wav2Lip & 50.86 $\to$ 54.82 & 61.89 $\to$ 59.69 & 46.48 $\to$ 47.98 & 69.13 $\to$ 64.25 & 72.88 $\to$ 84.10 & --- & 67.60 $\to$ 64.05 \\
      \bottomrule
    \end{tabular}%
  }
\end{table*}

\begin{table*}[tp]
  \centering
  \caption{Within-category adaptation on Entire Face Synthesis (EFS). Cells show scores before and after adaptation.}
  \label{tab:supp_efs}
  \vspace{-0.3cm}

  \resizebox{\textwidth}{!}{%
    \begin{tabular}{l|*{6}{c}|c}
      \toprule
       & \multicolumn{6}{|c|}{Source models} & \\
      \cmidrule(lr){2-7}
      Target & DiT-XL/2 & RDDM & StyleGAN-XL & VQGAN & StyleGAN3 & DDPM & Vanilla $\to$ Ours \\
      \midrule
      DiT-XL/2 & --- & 50.85 $\to$ 51.77 & 50.25 $\to$ 58.10 & 44.18 $\to$ 55.29 & 49.70 $\to$ 56.33 & 45.56 $\to$ 56.48 & 46.67 $\to$ 60.77 \\
      RDDM & 76.20 $\to$ 97.06 & --- & 49.82 $\to$ 93.11 & 99.73 $\to$ 99.91 & 46.83 $\to$ 86.22 & 99.71 $\to$ 99.86 & 96.02 $\to$ 100 \\
      StyleGAN-XL & 39.33 $\to$ 70.26 & 68.39 $\to$ 77.68 & --- & 44.80 $\to$ 52.69 & 70.21 $\to$ 90.44 & 46.83 $\to$ 47.84 & 57.46 $\to$ 95.30 \\
      VQGAN & 90.91 $\to$ 98.88 & 72.45 $\to$ 79.09 & 70.33 $\to$ 87.90 & --- & 73.28 $\to$ 95.30 & 61.73 $\to$ 89.14 & 93.16 $\to$ 99.87 \\
      StyleGAN3 & 61.64 $\to$ 95.85 & 11.07 $\to$ 72.07 & 55.95 $\to$ 86.94 & 63.14 $\to$ 90.68 & --- & 40.75 $\to$ 45.98 & 57.06 $\to$ 99.33 \\
      DDPM & 93.34 $\to$ 99.16 & 51.53 $\to$ 72.82 & 61.37 $\to$ 91.54 & 46.60 $\to$ 86.69 & 52.23 $\to$ 87.95 & --- & 93.56 $\to$ 98.73 \\
      \bottomrule
    \end{tabular}%
  }
\end{table*}

\section{More Results and Analysis}
\subsection{Detailed Results}
\noindent\textbf{High Performance Retention.}
Tables~\ref{tab:supp_fs} and~\ref{tab:supp_FR} show that the ensemble retains an AUC of 98.50\% to 100\% on SimSwap, MobileSwap, BlendFace, and the four FR targets other than the two audio-driven methods. On these targets, collaborative distillation also improves the weaker source models.

\smallskip
\noindent\textbf{Substantial Performance Gains.}
Table~\ref{tab:supp_efs} shows the largest gains in the Entire Face Synthesis category, where the average ensemble AUC increases from 73.99\% to 92.33\%, with StyleGAN3 jumping from 57.06\% to 99.33\% and StyleGAN-XL rising from 57.46\% to 95.30\%. Within Face Swapping, e4s also improves from 81.53\% to 91.80\%. From the adaptation results, we observe that for EFS targets, having just a few source models with moderate discriminability is sufficient to guide adaptation toward positive convergence. 

For instance, when StyleGAN-XL serves as the target domain, StyleGAN3 and RDDM start with initial AUCs of only 70.21\% and 68.39\%, respectively, yet they are enough to steer the adaptation process in the right direction. This positive feedback can even reverse source models with inverted predictions, lifting RDDM on StyleGAN3 sharply from 11.07\% to 72.07\%. The final ensemble results on StyleGAN3 and StyleGAN-XL reach 99.33\% and 95.30\%, surpassing the best adapted single models by 3.48\% and 4.86\%, respectively. This may be attributed to the fact that EFS forgery cues are distributed across the entire image. Through the interaction between collaborative distillation and iterative adaptation, our method integrates self-learning with multi-source auxiliary learning, preserving the specificity of each source model while fully exploiting complementary information across source models for further improvement. 

On e4s, which exhibits both regional composition boundaries and fine-grained texture cues covering a broad facial area, all five source models retain effective initial discriminability with AUCs ranging from 62.09\% to 80.69\%. During collaborative distillation and iterative adaptation, consensus from stronger source models continually lifts weaker source models, leading to simultaneous gains across all five models, with SimSwap jumping from 62.09\% to 76.70\% and InSwapper rising from 70.69\% to 84.08\%. The final five-source ensemble reaches 91.80\%, outperforming the strongest adapted single expert, BlendFace. By contrast, DiT-XL/2 presents a different case.Although all five source models improve after adaptation, the Transformer-based generation architecture of DiT differs substantially from the convolutional U-Net and GAN architectures underlying the source domains, potentially accounting for the near-random initial performance of all five source detectors, with AUCs ranging from 44.18\% to 50.85\% and thus limited effective positive guidance. Even so, the interaction within our method still brings a substantial improvement to the fused result.

\begin{table*}[tp]
  \centering
  \caption{Cross-category adaptation. Cells show scores before and after adaptation.}
  \vspace{-0.3cm}
  \label{tab:supp_cross_type}

  \resizebox{\textwidth}{!}{%
    \begin{tabular}{l|*{6}{c}|c}
      \toprule
       & \multicolumn{6}{|c|}{Source models} & \\
      \cmidrule(lr){2-7}
      Target & InSwapper & e4s & StyleGAN-XL & DiT-XL/2 & FOMM & Wav2Lip & Vanilla $\to$ Ours \\
      \midrule
      InSwapper & --- & 60.92 $\to$ 62.52 & 50.12 $\to$ 49.83 & 59.42 $\to$ 63.68 & 67.33 $\to$ 65.56 & 70.64 $\to$ 76.64 & 79.41 $\to$ 79.87 \\
      e4s & 70.69 $\to$ 85.95 & --- & 65.66 $\to$ 62.10 & 74.51 $\to$ 83.36 & 69.88 $\to$ 72.13 & 58.59 $\to$ 67.41 & 76.25 $\to$ 93.74 \\
      StyleGAN-XL & 83.56 $\to$ 97.51 & 38.95 $\to$ 54.44 & --- & 39.33 $\to$ 68.91 & 87.98 $\to$ 98.39 & 40.36 $\to$ 33.37 & 76.92 $\to$ 94.77 \\
      DiT-XL/2 & 51.88 $\to$ 47.83 & 52.63 $\to$ 54.84 & 50.25 $\to$ 49.14 & --- & 54.45 $\to$ 53.65 & 66.31 $\to$ 72.61 & 64.01 $\to$ 63.53 \\
      FOMM & 95.52 $\to$ 99.25 & 83.46 $\to$ 88.07 & 50.87 $\to$ 96.32 & 69.06 $\to$ 81.15 & --- & 61.41 $\to$ 71.93 & 93.44 $\to$ 98.20 \\
      Wav2Lip & 71.26 $\to$ 74.74 & 54.07 $\to$ 53.69 & 47.73 $\to$ 44.05 & 62.38 $\to$ 68.57 & 50.86 $\to$ 53.70 & --- & 65.86 $\to$ 69.21 \\
      \bottomrule
    \end{tabular}%
  }
\end{table*}

\smallskip
\noindent\textbf{Cross-Category Adaptation.}
Comparing within-category and cross-category adaptation results further supports the above analysis. As shown in table~\ref{tab:supp_cross_type}, when the overly homogeneous FR source models are replaced by cross-category source models including InSwapper and DiT-XL/2, Wav2Lip turns its decline into a gain, raising its ensemble AUC from 65.86\% to 69.21\%. This indicates that alleviating capability imbalance across source models and increasing source knowledge diversity is an effective option for enhancing adaptability on challenging domains. Despite a lower initial baseline of 76.25\%, e4s still reaches 93.74\%, surpassing its within-category adaptation result of 91.80\%. StyleGAN-XL achieves 94.77\%, largely benefiting from cross-category transfer from FOMM and InSwapper, which climb from 87.98\% to 98.39\% and from 83.56\% to 97.51\%, respectively. Meanwhile, even without similar FR source models, FOMM still adapts steadily from 93.44\% to 98.20\%. This shows that even though individual source models may be trained on different forgery families with distinct artifacts, our method can still extract useful complementary knowledge from these disparate source models. When DiT-XL/2 serves as the target domain, it still suffers from severe distribution shifts between the source and target domains. Although the initial average performance of the source models is higher and the Wav2Lip source improves from 66.31\% to 72.61\% after adaptation, this still cannot offset the drag from the other three weak source models, leaving the final result slightly lower than the vanilla baseline.

\begin{figure*}[!ht]
    \centering
    \includegraphics[width=\textwidth]{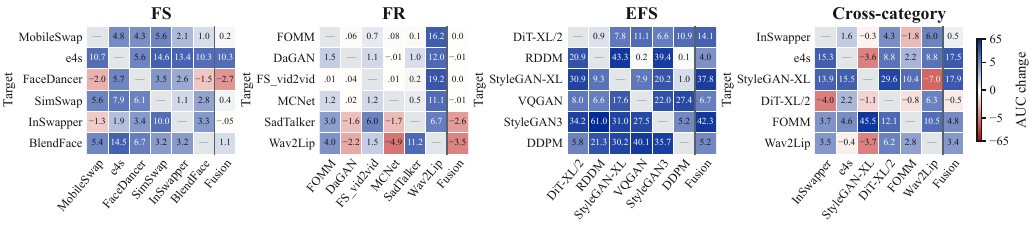}
    \vspace{-0.8cm}
    \caption{Heat map of AUC changes before and after adaptation for all adaptation tasks. A deeper shade of blue indicates a higher adaptation gain, while a deeper shade of red indicates more severe negative transfer.}
    \label{fig:auc_change}
    \vspace{-0.4cm}
\end{figure*}

\subsection{Phenomenon of Negative Transfer}
As shown in Figure~\ref{fig:auc_change}, Wav2Lip, FaceDancer, and SadTalker experience minor performance drops, indicating that degradation is not necessarily confined to hard domains with low baselines. Notably, on all three targets where the ensemble AUC declines, the arithmetic mean of the five adapted source models actually increases, moving from 60.25\% to 62.17\% on Wav2Lip, from 65.46\% to 67.13\% on FaceDancer, and from 82.50\% to 84.96\% on SadTalker. This suggests that the performance drop does not stem from a collapse of our adaptation method itself. Rather, we conjecture that it is caused by large distribution discrepancies between the target and source domains combined with an imbalance in capability among the source models, leading stronger source models to be suppressed by weaker ones. For example, on the audio-driven target Wav2Lip, the source pool consists of the four non-audio-driven FR source models and one audio-driven source model, SadTalker. Although SadTalker improves from 72.88\% to 84.10\%, it is inevitably outweighed in fusion by the other four mismatched source models. How to reduce the negative influence of weak source models requires further investigation.

\subsection{Distillation Losses and Multi-source Distillation Weight}
\begin{table*}[tp]
  \centering
  \caption{Effects of $\mathcal{L}_{\mathrm{sd}}$ and $\mathcal{L}_{\mathrm{md}}$. The first row uses the initial equal-weight source ensemble.}
  \label{tab:supp_loss_ablation}
    \vspace{-0.3cm}
  \setlength{\tabcolsep}{3pt}
  \renewcommand{\arraystretch}{1.12}

  \resizebox{\textwidth}{!}{%
  \begin{tabular}{cc|*{7}{c}|*{7}{c}|*{7}{c}|c}
    \toprule
     &  & \multicolumn{7}{|c}{Face Swap (FS)} & \multicolumn{7}{|c}{Face Reenactment (FR)} & \multicolumn{7}{|c|}{Entire Face Synthesis (EFS)} & \\
    \cmidrule(lr){3-9}\cmidrule(lr){10-16}\cmidrule(lr){17-23}

    $\mathcal{L}_{\mathrm{sd}}$ & $\mathcal{L}_{\mathrm{md}}$ &
    \rotatebox{90}{$\rightarrow$MobileSwap} & \rotatebox{90}{$\rightarrow$e4s} & \rotatebox{90}{$\rightarrow$FaceDancer} & \rotatebox{90}{$\rightarrow$SimSwap} & \rotatebox{90}{$\rightarrow$InSwapper} & \rotatebox{90}{$\rightarrow$BlendFace} & Avg. &
    \rotatebox{90}{$\rightarrow$FOMM} & \rotatebox{90}{$\rightarrow$DaGAN} & \rotatebox{90}{$\rightarrow$FS\_vid2vid} & \rotatebox{90}{$\rightarrow$MCNet} & \rotatebox{90}{$\rightarrow$SadTalker} & \rotatebox{90}{$\rightarrow$Wav2Lip} & Avg. &
    \rotatebox{90}{$\rightarrow$DiT-XL/2} & \rotatebox{90}{$\rightarrow$RDDM} & \rotatebox{90}{$\rightarrow$StyleGAN-XL} & \rotatebox{90}{$\rightarrow$VQGAN} & \rotatebox{90}{$\rightarrow$StyleGAN3} & \rotatebox{90}{$\rightarrow$DDPM} & Avg. &
    18 Avg. \\
    \midrule
     &  &
    98.27 & 81.53 & 74.82 & 98.50 & 84.85 & 97.93 & 89.32 &
    100 & 99.93 & 100 & 99.95 & 95.96 & 67.60 & 93.91 &
    46.67 & 96.02 & 57.46 & 93.16 & 57.06 & 93.56 & 73.99 &
    85.74 \\
    $\checkmark$ &  &
    99.12 & 91.63 & 71.97 & 99.33 & 85.17 & 99.13 & 91.06 &
    100 & 99.95 & 100 & 99.94 & 93.46 & 64.04 & 92.90 &
    61.31 & 100 & 85.39 & 99.83 & 95.84 & 99.78 & 90.36 &
    91.44 \\
    $\checkmark$ & $\checkmark$ &
    98.50 & 91.80 & 72.13 & 98.86 & 84.80 & 98.98 & 90.85 &
    100 & 99.92 & 100 & 99.94 & 93.39 & 64.05 & 92.88 &
    60.77 & 100 & 95.30 & 99.87 & 99.33 & 98.73 & 92.33 &
    92.02 \\
    \bottomrule
  \end{tabular}%
  }
\end{table*}

\begin{table*}[tp]
  \centering
  \caption{Sensitivity to the multi-source distillation weight $\lambda_1$.}
  \label{tab:supp_loss_mkd_weight}
      \vspace{-0.3cm}

  \resizebox{\textwidth}{!}{%
  \begin{tabular}{l|*{7}{c}|*{7}{c}|*{7}{c}|c}
    \toprule
     & \multicolumn{7}{|c}{Face Swap (FS)} & \multicolumn{7}{|c}{Face Reenactment (FR)} & \multicolumn{7}{|c|}{Entire Face Synthesis (EFS)} & \\
    \cmidrule(lr){2-8}\cmidrule(lr){9-15}\cmidrule(lr){16-22}

    $\lambda_1$ &
    \rotatebox{90}{$\rightarrow$MobileSwap} & \rotatebox{90}{$\rightarrow$e4s} & \rotatebox{90}{$\rightarrow$FaceDancer} & \rotatebox{90}{$\rightarrow$SimSwap} & \rotatebox{90}{$\rightarrow$InSwapper} & \rotatebox{90}{$\rightarrow$BlendFace} & Avg. &
    \rotatebox{90}{$\rightarrow$FOMM} & \rotatebox{90}{$\rightarrow$DaGAN} & \rotatebox{90}{$\rightarrow$FS\_vid2vid} & \rotatebox{90}{$\rightarrow$MCNet} & \rotatebox{90}{$\rightarrow$SadTalker} & \rotatebox{90}{$\rightarrow$Wav2Lip} & Avg. &
    \rotatebox{90}{$\rightarrow$DiT-XL/2} & \rotatebox{90}{$\rightarrow$RDDM} & \rotatebox{90}{$\rightarrow$StyleGAN-XL} & \rotatebox{90}{$\rightarrow$VQGAN} & \rotatebox{90}{$\rightarrow$StyleGAN3} & \rotatebox{90}{$\rightarrow$DDPM} & Avg. &
    18 Avg. \\
    \midrule
    0.05 &
    98.71 & 91.72 & 71.82 & 99.04 & 84.90 & 99.01 & 90.86 &
    100 & 99.95 & 100 & 99.95 & 93.29 & 63.66 & 92.81 &
    61.53 & 100 & 93.91 & 99.85 & 99.30 & 98.97 & 92.26 &
    91.98 \\
    0.10 (Ours) &
    98.50 & 91.80 & 72.13 & 98.86 & 84.80 & 98.98 & 90.85 &
    100 & 99.92 & 100 & 99.94 & 93.39 & 64.05 & 92.88 &
    60.77 & 100 & 95.30 & 99.87 & 99.33 & 98.73 & 92.33 &
    92.02 \\
    0.30 &
    97.89 & 92.33 & 72.41 & 98.62 & 84.84 & 98.77 & 90.81 &
    100 & 99.91 & 100 & 99.95 & 92.91 & 64.61 & 92.90 &
    59.28 & 99.99 & 83.02 & 99.88 & 98.84 & 98.83 & 89.97 &
    91.23 \\
    0.50 &
    97.91 & 92.53 & 72.98 & 98.47 & 85.59 & 98.64 & 91.02 &
    100 & 99.93 & 100 & 99.95 & 92.65 & 64.31 & 92.81 &
    59.44 & 100 & 86.92 & 99.90 & 96.81 & 98.98 & 90.34 &
    91.39 \\
    1 &
    98.27 & 93.22 & 73.54 & 98.55 & 86.37 & 98.56 & 91.42 &
    100 & 99.93 & 100 & 99.96 & 92.38 & 63.01 & 92.55 &
    58.55 & 100 & 73.51 & 99.94 & 93.24 & 99.42 & 87.44 &
    90.47 \\
    \bottomrule
  \end{tabular}%
  }
\end{table*}

Tables~\ref{tab:supp_loss_ablation} and~\ref{tab:supp_loss_mkd_weight} examine the individual roles of the single-source self-distillation loss and the multi-source collaborative distillation loss, as well as the weight of multi-source distillation.
From the results, using the single-source self-distillation loss alone already brings a substantial improvement, indicating that the self-evolution of source models on the target domain constitutes the primary source of adaptation gain, which benefits from appropriate self-distillation design choices.
Adding the multi-source collaborative distillation loss on top of this further raises the overall mean.

The sensitivity experiments on the weight $\lambda_1$ further reveal different responses of the three forgery categories to the strength of multi-source distillation.
Overall performance is most stable when $\lambda_1$ lies between 0.05 and 0.10, reaching 91.98\% and 92.02\%, respectively, whereas further enlarging the weight causes the mean to drop back to 90.47\%.
Specifically, increasing $\lambda_1$ is actually beneficial for the FS category, whose mean rises steadily from 90.85\% to 91.42\%, with improvements on e4s, InSwapper, and FaceDancer.
Although the methods in the FS category differ in how forgeries are generated, they share the same underlying generation paradigm, performing identity replacement and local blending only within the core facial region while preserving the target subject's background, pose, and expression (\blue{[4]}, \blue{[29]} in the main paper).
Our results also show that regardless of which method serves as the target domain, the initial test AUC of all other source models on the target domain remains above 60\%, without the completely inverted predictions seen among some source models in the EFS category.
Taking FaceDancer as an example, the initial average AUC of its source models is not high, yet strong sources with higher transferability still exist. Strengthening multi-source distillation helps weak sources that already have a certain level of discriminability improve further.

The EFS category shows the exact opposite trend. As $\lambda_1$ increases to 1, its mean drops from 92.33\% to 87.44\%, and StyleGAN-XL falls even further to 73.51\%.
Entire Face Synthesis exhibits the largest distribution shift between source and target domains among the three categories, with some sources even suffering from prediction inversion. An overly large multi-source weight forces strong models to compromise toward a noisy consensus while weakening their individual specificity.
We therefore set $\lambda_1$ to 0.10 by default, trading a slight 0.21\% concession on FS relative to pure self-distillation for substantial gains on challenging domains.
At the same time, mild negative transfer on SadTalker and Wav2Lip in FR already occurs during the pure self-distillation stage, lowering the FR mean from 93.91\% to 92.90\%, and adding the multi-source loss does not reverse this decline.

\subsection{Design Choices in Self-distillation}
\begin{table*}[tp]
  \centering
  \caption{Analysis of self-distillation design choices. TC denotes teacher centering, WS weak--strong views, FBN frozen batch-normalization statistics, and ST symmetric temperatures.}
  \label{tab:supp_self_kd_ablation}
    \vspace{-0.3cm}
  \setlength{\tabcolsep}{3pt}
  \renewcommand{\arraystretch}{1.12}

  \resizebox{\textwidth}{!}{%
  \begin{tabular}{cccc|*{7}{c}|*{7}{c}|*{7}{c}|c}
    \toprule
     &  &  &  & \multicolumn{7}{|c}{Face Swap (FS)} & \multicolumn{7}{|c}{Face Reenactment (FR)} & \multicolumn{7}{|c|}{Entire Face Synthesis (EFS)} & \\
    \cmidrule(lr){5-11}\cmidrule(lr){12-18}\cmidrule(lr){19-25}

    TC & WS & FBN & ST &
    \rotatebox{90}{$\rightarrow$MobileSwap} & \rotatebox{90}{$\rightarrow$e4s} & \rotatebox{90}{$\rightarrow$FaceDancer} & \rotatebox{90}{$\rightarrow$SimSwap} & \rotatebox{90}{$\rightarrow$InSwapper} & \rotatebox{90}{$\rightarrow$BlendFace} & Avg. &
    \rotatebox{90}{$\rightarrow$FOMM} & \rotatebox{90}{$\rightarrow$DaGAN} & \rotatebox{90}{$\rightarrow$FS\_vid2vid} & \rotatebox{90}{$\rightarrow$MCNet} & \rotatebox{90}{$\rightarrow$SadTalker} & \rotatebox{90}{$\rightarrow$Wav2Lip} & Avg. &
    \rotatebox{90}{$\rightarrow$DiT-XL/2} & \rotatebox{90}{$\rightarrow$RDDM} & \rotatebox{90}{$\rightarrow$StyleGAN-XL} & \rotatebox{90}{$\rightarrow$VQGAN} & \rotatebox{90}{$\rightarrow$StyleGAN3} & \rotatebox{90}{$\rightarrow$DDPM} & Avg. &
    18 Avg. \\
    \midrule
     & $\checkmark$ & $\checkmark$ & $\checkmark$ &
    99.32 & 83.42 & 62.87 & 99.16 & 86.52 & 99.18 & 88.41 &
    100 & 99.92 & 100 & 99.97 & 92.73 & 62.24 & 92.48 &
    43.98 & 99.87 & 42.96 & 99.40 & 71.37 & 99.75 & 76.22 &
    85.70 \\
    $\checkmark$ &  & $\checkmark$ & $\checkmark$ &
    97.03 & 84.98 & 74.05 & 97.01 & 83.11 & 97.09 & 88.88 &
    100 & 99.96 & 100 & 99.95 & 94.10 & 66.34 & 93.39 &
    62.46 & 99.98 & 79.31 & 97.74 & 83.15 & 96.31 & 86.49 &
    89.59 \\
    $\checkmark$ & $\checkmark$ &  & $\checkmark$ &
    99.36 & 93.14 & 80.81 & 99.22 & 90.70 & 99.10 & 93.72 &
    100 & 99.95 & 100 & 99.95 & 93.64 & 66.28 & 93.30 &
    58.78 & 100 & 78.40 & 99.92 & 77.28 & 98.09 & 85.41 &
    90.81 \\
    $\checkmark$ & $\checkmark$ & $\checkmark$ &  &
    98.27 & 91.21 & 71.28 & 98.77 & 84.43 & 98.85 & 90.47 &
    100 & 99.94 & 100 & 99.96 & 93.44 & 65.31 & 93.11 &
    59.34 & 100 & 83.19 & 99.78 & 88.16 & 98.64 & 88.19 &
    90.59 \\
    $\checkmark$ & $\checkmark$ & $\checkmark$ & $\checkmark$ &
    98.50 & 91.80 & 72.13 & 98.86 & 84.80 & 98.98 & 90.85 &
    100 & 99.92 & 100 & 99.94 & 93.39 & 64.05 & 92.88 &
    60.77 & 100 & 95.30 & 99.87 & 99.33 & 98.73 & 92.33 &
    92.02 \\
    \bottomrule
  \end{tabular}%
  }
\end{table*}

Table~\ref{tab:supp_self_kd_ablation} breaks down four design choices within the self-distillation module, including teacher centering, weak--strong view consistency, frozen batch normalization statistics, and symmetric temperatures, denoted as TC, WS, FBN, and ST, respectively.
Experiments show that all four components contribute positively to overall performance, with TC playing the most fundamental role.
Removing TC causes the 18-domain mean to drop sharply by 6.32\% to 85.70\%, nearly erasing all adaptation gains, while the EFS mean falls to 76.22\% and StyleGAN-XL and DiT-XL/2 even drop to 42.96\% and 43.98\%.
Under unlabeled settings with severe domain shifts, the absence of centering calibration easily drives teacher predictions toward a single class and triggers collapse.
Removing WS and ST also lowers the overall mean by 2.43\% and 1.43\%, respectively, indicating that strongly perturbed views help learn more invariant target features, while symmetric temperatures facilitate distribution alignment on difficult EFS targets.

By contrast, FBN exhibits clear divergence across categories.
Disabling FBN causes the EFS mean to drop sharply by 6.92\% to 85.41\%, with StyleGAN3 falling to 77.28\% and StyleGAN-XL to 78.40\%, which in turn pulls the overall mean down to 90.81\%.
At the same time, however, disabling FBN raises the FS mean by 2.87\% to 93.72\%, lifting FaceDancer and InSwapper to 80.81\% and 90.70\%, respectively, while the FR mean also increases slightly to 93.30\% with Wav2Lip reaching 66.28\%.

This divergence may stem from differences in forgery characteristics. FS and FR mainly involve modifications to local facial regions or specific facial attributes and typically retain much of the original image content, so their overall feature distributions may still share some similarity with real faces.
The target domain may introduce additional feature distribution shifts due to differences in lighting, color, or compression. Allowing BN statistics to update can help the model adapt to these additional distribution shifts, thereby reducing interference from forgery-irrelevant information during adaptation~\cite{schneider2020covariate}.
In contrast, EFS synthesizes entire faces through generative models, resulting in relatively pronounced statistical feature differences between fake and real images that form important discriminative cues for distinguishing the two~\cite{frank2020frequency}.
Although updating BN statistics online can facilitate target-domain adaptation, it also weakens the discriminative ability learned by source models prior to adaptation. Therefore, retaining source-domain BN statistics under EFS settings may better preserve the model's sensitivity to global generation artifacts.
Keeping FBN by default is thus a trade-off that balances overall performance with the adaptation ceiling on EFS, yet it also exposes the limitation of a fixed normalization strategy on localized manipulation domains, making an adaptive normalization mechanism that accounts for the spatial scope of forgeries necessary to explore in the future.

\subsection{Iterative Adaptation Strategies and Reliability Estimation Measures}
\begin{table*}[tb]
  \centering
  \caption{Analysis of iterative adaptation strategies. Joint optimizes all five models simultaneously with a single loss, with aggregation weights and fused soft targets refreshed after every batch. Fixed weights retains the initial aggregation weights throughout adaptation. Per-model and Per-batch refresh the weights and soft targets after each model update and each batch, respectively.}
  \label{tab:supp_refresh_strategy}
      \vspace{-0.3cm}
  \setlength{\tabcolsep}{3pt}
  \renewcommand{\arraystretch}{1.12}

  \resizebox{\textwidth}{!}{%
  \begin{tabular}{l|*{7}{c}|*{7}{c}|*{7}{c}|c}
    \toprule
     & \multicolumn{7}{|c}{Face Swap (FS)} & \multicolumn{7}{|c}{Face Reenactment (FR)} & \multicolumn{7}{|c|}{Entire Face Synthesis (EFS)} & \\
    \cmidrule(lr){2-8}\cmidrule(lr){9-15}\cmidrule(lr){16-22}

    Strategy &
    \rotatebox{90}{$\rightarrow$MobileSwap} & \rotatebox{90}{$\rightarrow$e4s} & \rotatebox{90}{$\rightarrow$FaceDancer} & \rotatebox{90}{$\rightarrow$SimSwap} & \rotatebox{90}{$\rightarrow$InSwapper} & \rotatebox{90}{$\rightarrow$BlendFace} & Avg. &
    \rotatebox{90}{$\rightarrow$FOMM} & \rotatebox{90}{$\rightarrow$DaGAN} & \rotatebox{90}{$\rightarrow$FS\_vid2vid} & \rotatebox{90}{$\rightarrow$MCNet} & \rotatebox{90}{$\rightarrow$SadTalker} & \rotatebox{90}{$\rightarrow$Wav2Lip} & Avg. &
    \rotatebox{90}{$\rightarrow$DiT-XL/2} & \rotatebox{90}{$\rightarrow$RDDM} & \rotatebox{90}{$\rightarrow$StyleGAN-XL} & \rotatebox{90}{$\rightarrow$VQGAN} & \rotatebox{90}{$\rightarrow$StyleGAN3} & \rotatebox{90}{$\rightarrow$DDPM} & Avg. &
    18 Avg. \\
    \midrule
    Joint &
    98.88 & 92.27 & 71.47 & 99.29 & 84.46 & 99.03 & 90.90 &
    100 & 99.92 & 100 & 99.94 & 94.69 & 64.11 & 93.11 &
    57.63 & 100 & 84.87 & 99.88 & 97.04 & 99.31 & 89.79 &
    91.27 \\
    Fixed weights &
    97.56 & 91.19 & 72.07 & 98.20 & 85.48 & 98.47 & 90.49 &
    100 & 99.93 & 100 & 99.95 & 91.43 & 63.49 & 92.47 &
    60.35 & 99.01 & 94.73 & 99.60 & 98.10 & 95.39 & 91.20 &
    91.39 \\
    Per-model &
    98.68 & 91.94 & 72.29 & 98.93 & 84.93 & 98.99 & 90.96 &
    100 & 99.95 & 100 & 99.93 & 93.28 & 63.60 & 92.79 &
    59.29 & 100 & 91.91 & 99.78 & 96.17 & 99.15 & 91.05 &
    91.60 \\
    Per-batch &
    99.03 & 91.61 & 71.77 & 99.07 & 85.17 & 98.94 & 90.93 &
    100 & 99.94 & 100 & 99.97 & 93.03 & 64.07 & 92.83 &
    60.28 & 100 & 93.97 & 99.86 & 95.50 & 99.43 & 91.51 &
    91.76 \\
    Per-round &
    98.50 & 91.80 & 72.13 & 98.86 & 84.80 & 98.98 & 90.85 &
    100 & 99.92 & 100 & 99.94 & 93.39 & 64.05 & 92.88 &
    60.77 & 100 & 95.30 & 99.87 & 99.33 & 98.73 & 92.33 &
    92.02 \\
    \bottomrule
  \end{tabular}%
  }
\end{table*}

\begin{table*}[tb]
  \centering
  \caption{Analysis of reliability estimation measures. $\mathcal{A}$ denotes prediction consensus, and $\mathcal{B}$ denotes feature prototype consistency.}
  \label{tab:supp_reliability_score_ablation}
  \vspace{-0.3cm}
  \setlength{\tabcolsep}{3pt}
  \renewcommand{\arraystretch}{1.12}

  \resizebox{\textwidth}{!}{%
  \begin{tabular}{l|*{7}{c}|*{7}{c}|*{7}{c}|c}
    \toprule
     & \multicolumn{7}{|c}{Face Swap (FS)} & \multicolumn{7}{|c}{Face Reenactment (FR)} & \multicolumn{7}{|c|}{Entire Face Synthesis (EFS)} & \\
    \cmidrule(lr){2-8}\cmidrule(lr){9-15}\cmidrule(lr){16-22}

    Reliability &
    \rotatebox{90}{$\rightarrow$MobileSwap} & \rotatebox{90}{$\rightarrow$e4s} & \rotatebox{90}{$\rightarrow$FaceDancer} & \rotatebox{90}{$\rightarrow$SimSwap} & \rotatebox{90}{$\rightarrow$InSwapper} & \rotatebox{90}{$\rightarrow$BlendFace} & Avg. &
    \rotatebox{90}{$\rightarrow$FOMM} & \rotatebox{90}{$\rightarrow$DaGAN} & \rotatebox{90}{$\rightarrow$FS\_vid2vid} & \rotatebox{90}{$\rightarrow$MCNet} & \rotatebox{90}{$\rightarrow$SadTalker} & \rotatebox{90}{$\rightarrow$Wav2Lip} & Avg. &
    \rotatebox{90}{$\rightarrow$DiT-XL/2} & \rotatebox{90}{$\rightarrow$RDDM} & \rotatebox{90}{$\rightarrow$StyleGAN-XL} & \rotatebox{90}{$\rightarrow$VQGAN} & \rotatebox{90}{$\rightarrow$StyleGAN3} & \rotatebox{90}{$\rightarrow$DDPM} & Avg. &
    18 Avg. \\
    \midrule
    Equal weights &
    98.67 & 91.58 & 72.19 & 98.88 & 84.95 & 98.90 & 90.86 &
    100 & 99.90 & 100 & 99.90 & 93.13 & 64.03 & 92.83 &
    60.61 & 100 & 92.17 & 99.85 & 97.24 & 99.49 & 91.56 &
    91.75 \\
    $+\mathcal{A}$ &
    98.57 & 91.66 & 72.13 & 98.90 & 84.66 & 98.90 & 90.80 &
    100 & 99.93 & 100 & 99.91 & 93.28 & 63.28 & 92.73 &
    61.00 & 100 & 92.12 & 99.83 & 96.48 & 99.42 & 91.48 &
    91.67 \\
    $+\mathcal{B}$ &
    98.68 & 91.89 & 72.21 & 98.89 & 84.77 & 98.95 & 90.90 &
    100 & 99.93 & 100 & 99.95 & 93.29 & 64.40 & 92.93 &
    60.58 & 99.99 & 93.57 & 99.84 & 99.54 & 99.39 & 92.15 &
    91.99 \\
    $+\mathcal{A}+\mathcal{B}$ &
    98.50 & 91.80 & 72.13 & 98.86 & 84.80 & 98.98 & 90.85 &
    100 & 99.92 & 100 & 99.94 & 93.39 & 64.05 & 92.88 &
    60.77 & 100 & 95.30 & 99.87 & 99.33 & 98.73 & 92.33 &
    92.02 \\
    \bottomrule
  \end{tabular}%
  }
\end{table*}

Table~\ref{tab:supp_refresh_strategy} compares five iterative adaptation strategies.
Although joint optimization holds a slight advantage on FR, with SadTalker reaching 94.69\%, it falls clearly behind on EFS where the source models exhibit larger initial disagreement, achieving a mean of only 89.79\% and dropping to 84.87\% on StyleGAN-XL.
This is mainly because the five models update simultaneously and overwrite the soft targets. These changes can introduce inaccurate supervision, reducing the reliability of the supervisory signals provided by the soft targets, which easily induces synchronized drift and may even cause the source models to homogenize prematurely.
Fixing the initial weights prevents the method from dynamically rewarding improving source models or suppressing degrading ones during adaptation, leading to clear declines on DDPM and SadTalker.
As for per-model and per-batch refreshing, overly high update frequencies easily introduce training fluctuations into the soft targets, making stable convergence harder on challenging domains such as StyleGAN3.
Our per-round refreshing allows the five source models to steadily complete a full round of evolution before recalibrating the weights and refreshing the soft targets together, achieving a better balance between stability and plasticity.

Table~\ref{tab:supp_reliability_score_ablation} further examines the roles of prediction consensus consistency $\mathcal{A}$ and feature prototype consistency $\mathcal{B}$ in reliability estimation.
An interesting phenomenon is that compared with equal-weight fusion at 91.75\%, introducing prediction consensus $\mathcal{A}$ alone slightly lowers the overall mean to 91.67\%, with small declines on both Wav2Lip and StyleGAN3.
This is because relying on prediction consensus amplifies majority bias when a majority of mismatched sources agree in the wrong direction.
By contrast, feature prototype consistency $\mathcal{B}$ alone raises the overall mean to 91.99\% and lifts the EFS mean from 91.56\% to 92.15\%, suggesting that prototype-based relative distance information can serve as a proxy for source model reliability to some extent.
Combining $\mathcal{A}$ and $\mathcal{B}$ achieves the highest overall mean of 92.02\% and yields an additional gain on StyleGAN-XL, raising it from 93.57\% to 95.30\%.
The two reliability metrics can provide complementary cues in certain cases, although the overall net gain over equal-weight fusion is modest, making them a low-cost optimization choice for adaptation.

\begin{figure}[!ht]
    \centering
    \includegraphics[width=\columnwidth]{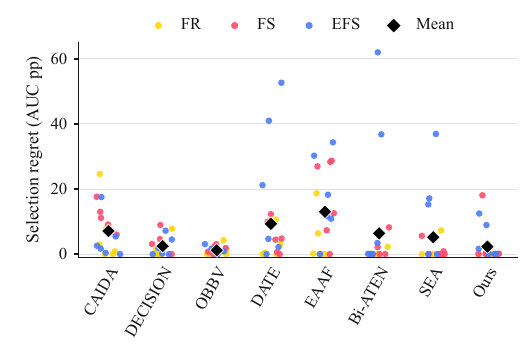}
    \vspace{-0.8cm}
    \caption{Each colored dot represents a target domain, with colors distinguishing the FR, FS, and EFS categories. Black diamonds indicate the mean across target domains. Lower values indicate that the selected source model is closer to the actual best model, resulting in a lower cost of incorrect source selection.}
    \vspace{-0.4cm}
    \label{fig:selection_regret}
\end{figure}

\begin{figure}[!ht]
    \centering
    \includegraphics[width=\columnwidth]{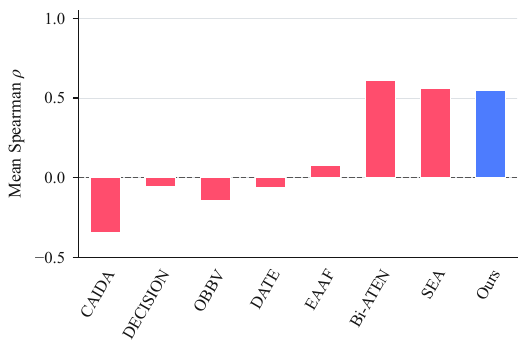}
    \vspace{-0.8cm}
    \caption{Each bar shows the mean correlation over target domains for which both rankings have nonzero variance. Higher values indicate closer agreement between the estimated source ranking and the ranking by target AUC.}
    \vspace{-0.2cm}
    \label{fig:spearman}
\end{figure}

We also analyze from two aspects whether our method and the baselines accurately estimate the importance of source models. Figure~\ref{fig:selection_regret} shows the AUC gap between the best source model selected by each method and the actual best source model, which also reflects to some extent the cost each method pays once an incorrect selection is made. Then, for each target domain, we rank the five source models according to their calculated weights and actual AUC values, respectively, and compute the Spearman correlation coefficient between the two rankings~\cite{spearman1961}. Figure~\ref{fig:spearman} shows the average Spearman correlation coefficient of each method. As can be seen, only SEA, Bi-ATEN, and our method rank the importance of the source models relatively accurately. However, in Figure~\ref{fig:selection_regret}, the maximum gaps for SEA and Bi-ATEN are around 40\% and 60\%, respectively, which brings sustained negative effects during adaptation and also shows that their estimation methods are not sufficiently stable. Our method is competitive in both aspects and achieves the best overall performance.

\subsection{Adaptation Using Various Backbones}
\begin{table*}[tb]
  \centering
  \caption{Analysis of adaptation using various backbones. Before uses initial equal-weight fusion and After uses the weighted fusion after three rounds.}
  \label{tab:supp_transferability_backbone}
  \setlength{\tabcolsep}{3pt}
  \renewcommand{\arraystretch}{1.12}

  \resizebox{\textwidth}{!}{%
  \begin{tabular}{ll|*{7}{c}|*{7}{c}|*{7}{c}|c}
    \toprule
     &  & \multicolumn{7}{|c}{Face Swap (FS)} & \multicolumn{7}{|c}{Face Reenactment (FR)} & \multicolumn{7}{|c|}{Entire Face Synthesis (EFS)} & \\
    \cmidrule(lr){3-9}\cmidrule(lr){10-16}\cmidrule(lr){17-23}

    Backbone & Stage &
    \rotatebox{90}{$\rightarrow$MobileSwap} & \rotatebox{90}{$\rightarrow$e4s} & \rotatebox{90}{$\rightarrow$FaceDancer} & \rotatebox{90}{$\rightarrow$SimSwap} & \rotatebox{90}{$\rightarrow$InSwapper} & \rotatebox{90}{$\rightarrow$BlendFace} & Avg. &
    \rotatebox{90}{$\rightarrow$FOMM} & \rotatebox{90}{$\rightarrow$DaGAN} & \rotatebox{90}{$\rightarrow$FS\_vid2vid} & \rotatebox{90}{$\rightarrow$MCNet} & \rotatebox{90}{$\rightarrow$SadTalker} & \rotatebox{90}{$\rightarrow$Wav2Lip} & Avg. &
    \rotatebox{90}{$\rightarrow$DiT-XL/2} & \rotatebox{90}{$\rightarrow$RDDM} & \rotatebox{90}{$\rightarrow$StyleGAN-XL} & \rotatebox{90}{$\rightarrow$VQGAN} & \rotatebox{90}{$\rightarrow$StyleGAN3} & \rotatebox{90}{$\rightarrow$DDPM} & Avg. &
    18 Avg. \\
    \midrule
    Xception & Before &
    98.27 & 81.53 & 74.82 & 98.50 & 84.85 & 97.93 & 89.32 &
    100 & 99.93 & 100 & 99.95 & 95.96 & 67.60 & 93.91 &
    46.67 & 96.02 & 57.46 & 93.16 & 57.06 & 93.56 & 73.99 &
    85.74 \\
     & After &
    98.50 & 91.80 & 72.13 & 98.86 & 84.80 & 98.98 & 90.85 &
    100 & 99.92 & 100 & 99.94 & 93.39 & 64.05 & 92.88 &
    60.77 & 100 & 95.30 & 99.87 & 99.33 & 98.73 & 92.33 &
    92.02 \\
    \midrule
    ResNet-50 & Before &
    96.20 & 77.16 & 72.71 & 97.63 & 74.06 & 95.94 & 85.62 &
    99.92 & 99.81 & 100 & 99.92 & 83.04 & 63.98 & 91.11 &
    49.52 & 92.23 & 68.83 & 88.92 & 46.19 & 94.50 & 73.36 &
    83.36 \\
     & After &
    98.54 & 88.06 & 76.65 & 98.42 & 78.02 & 98.12 & 89.64 &
    99.98 & 99.81 & 100 & 99.89 & 87.25 & 56.85 & 90.63 &
    51.33 & 100 & 74.70 & 98.23 & 72.02 & 98.26 & 82.43 &
    87.56 \\
    \midrule
    EfficientNet-B4 & Before &
    97.13 & 85.79 & 81.09 & 98.11 & 85.16 & 97.67 & 90.83 &
    100 & 99.99 & 100 & 99.98 & 96.43 & 77.16 & 95.59 &
    59.10 & 100 & 81.35 & 98.87 & 98.11 & 98.73 & 89.36 &
    91.93 \\
     & After &
    98.85 & 96.34 & 85.33 & 99.28 & 93.05 & 99.51 & 95.39 &
    100 & 99.98 & 100 & 99.99 & 96.25 & 75.25 & 95.25 &
    73.17 & 100 & 95.90 & 99.81 & 100 & 99.96 & 94.81 &
    95.15 \\
    \midrule
    Swin-Tiny & Before &
    74.40 & 62.16 & 55.19 & 83.83 & 64.14 & 81.50 & 70.20 &
    94.99 & 98.50 & 99.78 & 98.97 & 72.68 & 55.83 & 86.79 &
    47.23 & 82.89 & 61.33 & 56.78 & 69.08 & 72.90 & 65.03 &
    74.01 \\
     & After &
    87.69 & 76.90 & 58.19 & 89.81 & 71.60 & 81.13 & 77.55 &
    95.28 & 99.05 & 99.88 & 99.18 & 71.46 & 54.49 & 86.55 &
    56.10 & 95.49 & 67.07 & 83.30 & 84.11 & 51.10 & 72.86 &
    78.99 \\
    \midrule
    DeiT-Small/16 & Before &
    60.09 & 58.82 & 58.52 & 67.92 & 54.58 & 72.86 & 62.13 &
    90.46 & 98.35 & 99.35 & 98.50 & 62.72 & 53.15 & 83.75 &
    51.78 & 52.79 & 68.76 & 56.80 & 55.90 & 74.69 & 60.12 &
    68.67 \\
     & After &
    57.49 & 58.61 & 59.67 & 68.59 & 51.12 & 68.90 & 60.73 &
    90.95 & 98.89 & 99.52 & 99.03 & 64.27 & 57.07 & 84.95 &
    50.35 & 62.79 & 92.05 & 80.43 & 68.34 & 81.40 & 72.56 &
    72.75 \\
    \bottomrule
  \end{tabular}%
  }
\end{table*}

\begin{table*}[!htb]
  \centering
  \caption{Influence of the number of interaction rounds.}
  \label{tab:supp_number_of_rounds}
  \vspace{-0.3cm}
  \setlength{\tabcolsep}{3pt}
  \renewcommand{\arraystretch}{1.12}

  \resizebox{\textwidth}{!}{%
  \begin{tabular}{l|*{7}{c}|*{7}{c}|*{7}{c}|c}
    \toprule
     & \multicolumn{7}{|c}{Face Swap (FS)} & \multicolumn{7}{|c}{Face Reenactment (FR)} & \multicolumn{7}{|c|}{Entire Face Synthesis (EFS)} & \\
    \cmidrule(lr){2-8}\cmidrule(lr){9-15}\cmidrule(lr){16-22}

    Rounds &
    \rotatebox{90}{$\rightarrow$MobileSwap} & \rotatebox{90}{$\rightarrow$e4s} & \rotatebox{90}{$\rightarrow$FaceDancer} & \rotatebox{90}{$\rightarrow$SimSwap} & \rotatebox{90}{$\rightarrow$InSwapper} & \rotatebox{90}{$\rightarrow$BlendFace} & Avg. &
    \rotatebox{90}{$\rightarrow$FOMM} & \rotatebox{90}{$\rightarrow$DaGAN} & \rotatebox{90}{$\rightarrow$FS\_vid2vid} & \rotatebox{90}{$\rightarrow$MCNet} & \rotatebox{90}{$\rightarrow$SadTalker} & \rotatebox{90}{$\rightarrow$Wav2Lip} & Avg. &
    \rotatebox{90}{$\rightarrow$DiT-XL/2} & \rotatebox{90}{$\rightarrow$RDDM} & \rotatebox{90}{$\rightarrow$StyleGAN-XL} & \rotatebox{90}{$\rightarrow$VQGAN} & \rotatebox{90}{$\rightarrow$StyleGAN3} & \rotatebox{90}{$\rightarrow$DDPM} & Avg. &
    18 Avg. \\
    \midrule
    1 &
    98.71 & 88.44 & 72.68 & 99.02 & 84.79 & 98.82 & 90.41 &
    100 & 99.95 & 100 & 99.98 & 94.94 & 61.23 & 92.68 &
    59.89 & 100 & 91.75 & 99.64 & 96.93 & 98.18 & 91.06 &
    91.39 \\
    2 &
    98.13 & 89.58 & 71.57 & 98.40 & 84.80 & 98.79 & 90.21 &
    100 & 99.93 & 100 & 99.96 & 93.85 & 61.85 & 92.60 &
    60.39 & 100 & 92.90 & 99.76 & 98.70 & 98.80 & 91.76 &
    91.52 \\
    3 (Ours) &
    98.50 & 91.80 & 72.13 & 98.86 & 84.80 & 98.98 & 90.85 &
    100 & 99.92 & 100 & 99.94 & 93.39 & 64.05 & 92.88 &
    60.77 & 100 & 95.30 & 99.87 & 99.33 & 98.73 & 92.33 &
    92.02 \\
    4 &
    98.70 & 92.15 & 72.14 & 98.97 & 85.67 & 99.04 & 91.11 &
    100 & 99.95 & 100 & 99.94 & 93.25 & 63.36 & 92.75 &
    60.35 & 100 & 94.79 & 99.93 & 99.66 & 98.87 & 92.27 &
    92.04 \\
    5 &
    98.89 & 92.78 & 72.40 & 99.01 & 85.72 & 98.98 & 91.30 &
    100 & 99.95 & 100 & 99.94 & 93.29 & 65.05 & 93.04 &
    60.33 & 100 & 96.59 & 99.92 & 99.63 & 98.92 & 92.57 &
    92.30 \\
    6 &
    98.90 & 93.01 & 71.48 & 98.92 & 86.67 & 99.07 & 91.34 &
    100 & 99.94 & 100 & 99.94 & 93.10 & 64.46 & 92.91 &
    60.37 & 100 & 96.19 & 99.93 & 99.77 & 98.96 & 92.54 &
    92.26 \\
    7 &
    98.92 & 93.81 & 72.78 & 99.00 & 86.64 & 99.15 & 91.72 &
    100 & 99.94 & 100 & 99.94 & 93.80 & 65.10 & 93.13 &
    60.19 & 100 & 96.67 & 99.94 & 99.90 & 98.97 & 92.61 &
    92.49 \\
    8 &
    98.84 & 93.85 & 71.59 & 98.97 & 87.17 & 99.08 & 91.58 &
    100 & 99.95 & 100 & 99.94 & 93.86 & 65.05 & 93.13 &
    60.55 & 100 & 97.34 & 99.95 & 99.84 & 98.90 & 92.76 &
    92.49 \\
    9 &
    98.97 & 93.86 & 71.83 & 98.97 & 87.81 & 99.18 & 91.77 &
    100 & 99.95 & 100 & 99.92 & 94.34 & 65.58 & 93.30 &
    60.47 & 100 & 96.86 & 99.96 & 99.86 & 98.91 & 92.68 &
    92.58 \\
    10 &
    98.89 & 94.04 & 71.85 & 99.02 & 88.10 & 99.12 & 91.84 &
    100 & 99.95 & 100 & 99.93 & 93.73 & 65.94 & 93.26 &
    60.39 & 100 & 96.71 & 99.97 & 99.85 & 98.69 & 92.60 &
    92.57 \\
    \bottomrule
  \end{tabular}%
  }
\end{table*}

\begin{table*}[!htb]
  \centering
  \caption{Influence of the number of student epochs per round.}
  \label{tab:supp_number_of_epochs}
  \vspace{-0.3cm}
  \setlength{\tabcolsep}{3pt}
  \renewcommand{\arraystretch}{1.12}

  \resizebox{\textwidth}{!}{%
  \begin{tabular}{l|*{7}{c}|*{7}{c}|*{7}{c}|c}
    \toprule
     & \multicolumn{7}{|c}{Face Swap (FS)} & \multicolumn{7}{|c}{Face Reenactment (FR)} & \multicolumn{7}{|c|}{Entire Face Synthesis (EFS)} & \\
    \cmidrule(lr){2-8}\cmidrule(lr){9-15}\cmidrule(lr){16-22}

    Epochs &
    \rotatebox{90}{$\rightarrow$MobileSwap} & \rotatebox{90}{$\rightarrow$e4s} & \rotatebox{90}{$\rightarrow$FaceDancer} & \rotatebox{90}{$\rightarrow$SimSwap} & \rotatebox{90}{$\rightarrow$InSwapper} & \rotatebox{90}{$\rightarrow$BlendFace} & Avg. &
    \rotatebox{90}{$\rightarrow$FOMM} & \rotatebox{90}{$\rightarrow$DaGAN} & \rotatebox{90}{$\rightarrow$FS\_vid2vid} & \rotatebox{90}{$\rightarrow$MCNet} & \rotatebox{90}{$\rightarrow$SadTalker} & \rotatebox{90}{$\rightarrow$Wav2Lip} & Avg. &
    \rotatebox{90}{$\rightarrow$DiT-XL/2} & \rotatebox{90}{$\rightarrow$RDDM} & \rotatebox{90}{$\rightarrow$StyleGAN-XL} & \rotatebox{90}{$\rightarrow$VQGAN} & \rotatebox{90}{$\rightarrow$StyleGAN3} & \rotatebox{90}{$\rightarrow$DDPM} & Avg. &
    18 Avg. \\
    \midrule
    1 (Ours) &
    98.50 & 91.80 & 72.13 & 98.86 & 84.80 & 98.98 & 90.85 &
    100 & 99.92 & 100 & 99.94 & 93.39 & 64.05 & 92.88 &
    60.77 & 100 & 95.30 & 99.87 & 99.33 & 98.73 & 92.33 &
    92.02 \\
    2 &
    98.55 & 93.34 & 71.88 & 98.94 & 88.06 & 98.82 & 91.60 &
    100 & 99.94 & 100 & 99.93 & 92.27 & 65.05 & 92.86 &
    60.17 & 100 & 96.51 & 99.90 & 99.59 & 96.63 & 92.13 &
    92.20 \\
    3 &
    98.71 & 94.86 & 72.42 & 98.89 & 88.93 & 98.89 & 92.11 &
    100 & 99.91 & 100 & 99.91 & 92.21 & 66.15 & 93.03 &
    61.18 & 100 & 96.97 & 99.86 & 99.45 & 95.24 & 92.12 &
    92.42 \\
    4 &
    98.62 & 95.15 & 72.15 & 98.78 & 89.51 & 98.71 & 92.15 &
    100 & 99.87 & 100 & 99.89 & 92.17 & 66.34 & 93.04 &
    60.54 & 99.99 & 97.80 & 99.82 & 98.73 & 93.22 & 91.68 &
    92.29 \\
    5 &
    98.47 & 95.45 & 73.32 & 98.65 & 89.78 & 98.90 & 92.43 &
    100 & 99.84 & 100 & 99.89 & 92.04 & 66.47 & 93.04 &
    60.05 & 99.98 & 97.89 & 99.80 & 97.66 & 91.66 & 91.17 &
    92.21 \\
    \bottomrule
  \end{tabular}%
  }
\end{table*}

\begin{table*}[!htb]
  \centering
  \caption{Impact of the EMA coefficient $m_{\mathrm{t}}$ in self-distillation.}
  \label{tab:supp_ema_self_kd}
  \vspace{-0.3cm}
  \setlength{\tabcolsep}{3pt}
  \renewcommand{\arraystretch}{1.12}

  \resizebox{\textwidth}{!}{%
  \begin{tabular}{l|*{7}{c}|*{7}{c}|*{7}{c}|c}
    \toprule
     & \multicolumn{7}{|c}{Face Swap (FS)} & \multicolumn{7}{|c}{Face Reenactment (FR)} & \multicolumn{7}{|c|}{Entire Face Synthesis (EFS)} & \\
    \cmidrule(lr){2-8}\cmidrule(lr){9-15}\cmidrule(lr){16-22}

    $m_{\mathrm{t}}$ &
    \rotatebox{90}{$\rightarrow$MobileSwap} & \rotatebox{90}{$\rightarrow$e4s} & \rotatebox{90}{$\rightarrow$FaceDancer} & \rotatebox{90}{$\rightarrow$SimSwap} & \rotatebox{90}{$\rightarrow$InSwapper} & \rotatebox{90}{$\rightarrow$BlendFace} & Avg. &
    \rotatebox{90}{$\rightarrow$FOMM} & \rotatebox{90}{$\rightarrow$DaGAN} & \rotatebox{90}{$\rightarrow$FS\_vid2vid} & \rotatebox{90}{$\rightarrow$MCNet} & \rotatebox{90}{$\rightarrow$SadTalker} & \rotatebox{90}{$\rightarrow$Wav2Lip} & Avg. &
    \rotatebox{90}{$\rightarrow$DiT-XL/2} & \rotatebox{90}{$\rightarrow$RDDM} & \rotatebox{90}{$\rightarrow$StyleGAN-XL} & \rotatebox{90}{$\rightarrow$VQGAN} & \rotatebox{90}{$\rightarrow$StyleGAN3} & \rotatebox{90}{$\rightarrow$DDPM} & Avg. &
    18 Avg. \\
    \midrule
    0.990 &
    98.77 & 92.74 & 69.25 & 98.95 & 83.15 & 99.07 & 90.32 &
    100 & 99.91 & 100 & 99.96 & 92.17 & 63.45 & 92.58 &
    58.55 & 100 & 95.24 & 99.85 & 99.86 & 97.40 & 91.82 &
    91.57 \\
    0.993 &
    98.73 & 92.75 & 70.17 & 98.94 & 84.80 & 98.98 & 90.73 &
    100 & 99.90 & 100 & 99.93 & 92.82 & 63.09 & 92.62 &
    59.76 & 100 & 95.26 & 99.87 & 99.78 & 98.68 & 92.22 &
    91.86 \\
    0.996 (Ours) &
    98.50 & 91.80 & 72.13 & 98.86 & 84.80 & 98.98 & 90.85 &
    100 & 99.92 & 100 & 99.94 & 93.39 & 64.05 & 92.88 &
    60.77 & 100 & 95.30 & 99.87 & 99.33 & 98.73 & 92.33 &
    92.02 \\
    0.999 &
    98.54 & 89.78 & 74.70 & 98.58 & 85.13 & 98.76 & 90.91 &
    100 & 99.92 & 100 & 99.97 & 94.38 & 63.99 & 93.04 &
    61.85 & 100 & 88.52 & 99.48 & 95.04 & 98.46 & 90.56 &
    91.51 \\
    \bottomrule
  \end{tabular}%
  }
\end{table*}

Table~\ref{tab:supp_transferability_backbone} tests the architectural generality of our framework across three convolutional networks and two vision Transformers. We select backbones with comparable parameter counts to avoid unfairness in evaluating adaptation capability caused by large parameter sizes.
Overall, regardless of how the underlying backbone varies, our method consistently improves the 18-domain mean by 3.22\% to 6.28\%, with gains of 6.28\% on Xception, 4.98\% on Swin-Tiny, 4.20\% on ResNet-50, 4.08\% on DeiT-Small/16, and 3.22\% on EfficientNet-B4.
When switching to EfficientNet-B4 with stronger feature representations, the adapted 18-domain mean reaches 95.15\%, the FS mean rises from 90.83\% to 95.39\%, and the highly challenging target DiT-XL/2 also jumps from 59.10\% to 73.17\%.

Comparing results across different backbones also allows a more objective view of negative transfer on certain domains.
FaceDancer, which previously showed a slight drop on Xception, turns into positive improvements across all four remaining backbones, gaining 3.94\% on ResNet-50, 4.24\% on EfficientNet-B4, 3.00\% on Swin-Tiny, and 1.15\% on DeiT-Small/16.
SadTalker similarly gains 4.21\% on ResNet-50 and 1.55\% on DeiT-Small/16, while Wav2Lip also improves by 3.92\% on DeiT-Small/16.
This shows that performance drops on certain target domains are not a general flaw of the framework's optimization objective, but are also closely related to the backbone's initial sensitivity to specific forgery traces.
The results also expose the relatively weak detection capability of compact vision Transformers in cross-domain scenarios, possibly reflecting the limitations of patch-based feature representations in capturing fine-grained, high-frequency local artifacts in the target domain. Both Swin-Tiny and DeiT-Small/16 have markedly lower initial baselines than the convolutional networks, and adaptation fails to close this gap.
Moreover, none of the backbones solves the adaptation difficulty on challenging domains such as Wav2Lip, which requires further research in the future.

\subsection{Interaction Rounds, Student Epochs, and Teacher Momentum}
Tables~\ref{tab:supp_number_of_rounds}, \ref{tab:supp_number_of_epochs}, and~\ref{tab:supp_ema_self_kd} further investigate three hyperparameters.
As the number of interaction rounds increases from 1 to 10, the 18-domain mean exhibits an overall upward trend despite minor fluctuations, rising from 91.39\% at round 1 to 92.02\% at round 3 and reaching 92.58\% and 92.57\% at rounds 9 and 10, respectively, with even Wav2Lip showing an overall improvement from 61.23\% at round 1 to 65.94\% at round 10. This demonstrates that the framework has strong resistance to drift over prolonged unsupervised iterations. Selecting 3 rounds by default is mainly intended to balance efficiency and performance gain.

Comparing Table~\ref{tab:supp_number_of_rounds} with Table~\ref{tab:supp_number_of_epochs} reveals a key optimization pattern.
Although increasing the number of student epochs per round slightly pushes the overall mean up to 92.42\% at 3 epochs and steadily lifts FS from 90.85\% to 92.43\% and Wav2Lip from 64.05\% to 66.47\%, increasing it further causes the overall mean to decline.
Increasing the number of epochs causes continuous degradation on EFS from the outset, with the EFS mean falling from 92.33\% to 91.17\% and DDPM sliding steadily from 98.73\% to 91.66\%.
By contrast, under the setting of 10 interaction rounds with only 1 epoch per round, DDPM still stays steadily at 98.69\%.
We believe this is likewise caused by the differences between the FS and EFS categories. As analyzed earlier, FS source models exhibit much more stable initial discriminability on the target domain than EFS source models and can more readily provide effective supervision, which can be absorbed further as the number of epochs increases.
By contrast, EFS faces huge initial distribution shifts, and individual source models undergo large evolutionary changes during adaptation, so each source model must be updated promptly. Otherwise, uncorrected weak sources continue to drag down the adaptation process.
These results indicate that under our experimental protocol, iterative adaptation with more rounds and smaller steps generally outperforms isolated self-training with fewer rounds and larger steps.

The two extremes of the teacher momentum coefficient $m_{\mathrm{t}}$ in Table~\ref{tab:supp_ema_self_kd} also clearly reflect the trade-off between stability and plasticity.
When $m_{\mathrm{t}}$ is small at 0.990, the teacher model follows noisy early student updates too quickly, lowering performance on FaceDancer, InSwapper, SadTalker, and DiT-XL/2.
When $m_{\mathrm{t}}$ is too large at 0.999, the teacher model relies more heavily on prior knowledge from source pretraining. Although this alleviates negative transfer to some extent, raising FaceDancer to 74.70\%, SadTalker to 94.38\%, and DiT-XL/2 to 61.85\%, it also clearly restricts adaptation processes that require timely adjustments to decision boundaries, causing StyleGAN-XL to drop to 88.52\%, StyleGAN3 to 95.04\%, and e4s to 89.78\%.
Setting $m_{\mathrm{t}}$ to 0.996 by default accommodates both of these distinct adaptation needs and achieves the best overall balance.

{
    \small
    \putbib[main]
}

\clearpage
\end{bibunit}

\end{document}